%% file: arxiv.tex
\documentclass{article}
\usepackage{iclr2027_conference,times}

\input{math_commands.tex}

\usepackage{amsmath,amssymb,mathtools}
\usepackage{graphicx}
\usepackage{booktabs}
\usepackage{multirow}
\usepackage{array}
\usepackage{microtype}
\usepackage{xcolor}
\usepackage{url}
\usepackage{hyperref}
\usepackage{enumitem}
\usepackage{caption}
\usepackage{subcaption}
\usepackage{algorithm}
\usepackage{algpseudocode}
\usepackage{placeins}
\usepackage{xspace}
\usepackage{makecell}
\usepackage[normalem]{ulem}

\definecolor{oursblue}{HTML}{2457D6}
\definecolor{oursorange}{HTML}{E76F00}
\definecolor{oursgreen}{HTML}{16866B}
\definecolor{oursgray}{HTML}{5B677A}
\definecolor{oursred}{HTML}{C43D3D}

\newcommand{\system}{\textsc{Pulse}\xspace}

\title{\system: Unlocking Practical Image Compression on Single-Thread CPU}

\author{%
\parbox[t]{\dimexpr\textwidth-2\tabcolsep\relax}{\raggedright
\mbox{\textbf{Zhaoyang Jia}$^{1}$\thanks{This work was done when Zhaoyang Jia and Zihan Zheng were full-time interns at Microsoft Research Asia.}}\quad
\mbox{\textbf{Tianyu Zhang}$^{\ddagger}$}\quad
\mbox{\textbf{Zihan Zheng}$^{1}$\footnotemark[1]}\quad
\mbox{\textbf{Wenxuan Xie}$^{2}$}\quad
\mbox{\textbf{Jiahao Li}$^{2}$}\quad
\mbox{\textbf{Bin Li}$^{2}$}\quad
\mbox{\textbf{Houqiang Li}$^{1}$}\quad
\mbox{\textbf{Yan Lu}$^{2}$} \\[5pt]
\textnormal{$^{1}$University of Science and Technology of China} \\
\textnormal{$^{2}$Microsoft Research Asia \quad $^{\ddagger}$Independent Researcher}
}%
}

\iclrfinalcopy % Show authors and disable review line numbers.

\begin{document}
\maketitle
% Override the conference-publication header set by \maketitle.
\lhead{Preprint}
\begingroup
\renewcommand{\thefootnote}{}
\footnotetext[0]{Corresponding to:  \href{mailto:jzy_ustc@mail.ustc.edu.cn}{\texttt{jzy\_ustc@mail.ustc.edu.cn}} (Zhaoyang Jia).}
\endgroup

\input{sec/0_abstract}

\input{sec/1_intro}
\input{sec/2_architecture}
\input{sec/3_crossplatform}
\input{sec/4_agentic_evolution}
\input{sec/5_perceptual}
\input{sec/6_experiments}
\input{sec/7_landscape}
\input{sec/8_conclusion}

% Required disclosure, outside the main-text page budget.
% \clearpage
% \input{sec/9_ai_use}

\bibliography{references}
\bibliographystyle{iclr2027_conference}

\appendix
\input{sec/X_supp}

\end{document}

%% file: math_commands.tex
\usepackage{amsmath,amsfonts,bm}

\def\eqref#1{equation~\ref{#1}}
\def\1{\bm{1}}

\DeclareMathAlphabet{\mathsfit}{\encodingdefault}{\sfdefault}{m}{sl}
\SetMathAlphabet{\mathsfit}{bold}{\encodingdefault}{\sfdefault}{bx}{n}

%% file: sec/0_abstract.tex
\begin{abstract}

Despite recent progress in learned image compression, existing methods remain computationally expensive on resource-constrained hardware, particularly CPUs.
We introduce \system, a practical codec that enables (1) \textbf{low-latency decoding on diverse hardware platforms} with an ultra-low-complexity 5.2~kMAC/pixel neural receiver, and (2) \textbf{efficient bit-exact entropy coding} with an integer linear CDF predictor and a meta prior.
To recover compression performance under this tight budget, we introduce an agentic evolution process guided by heuristic probes that iteratively improves the architecture through human--LLM collaboration.
\system decodes a 1080p image in 126~ms on \textbf{a single CPU thread} while achieving compression performance comparable to HM.
After perceptual optimization, \system competes with larger perceptual codecs like MS-ILLM.
Codes are at \url{https://github.com/microsoft/GenCodec/tree/main/PULSE}.
\end{abstract}

%% file: sec/1_intro.tex
\begin{figure*}[h]
    \centering
    \includegraphics[width=\textwidth]{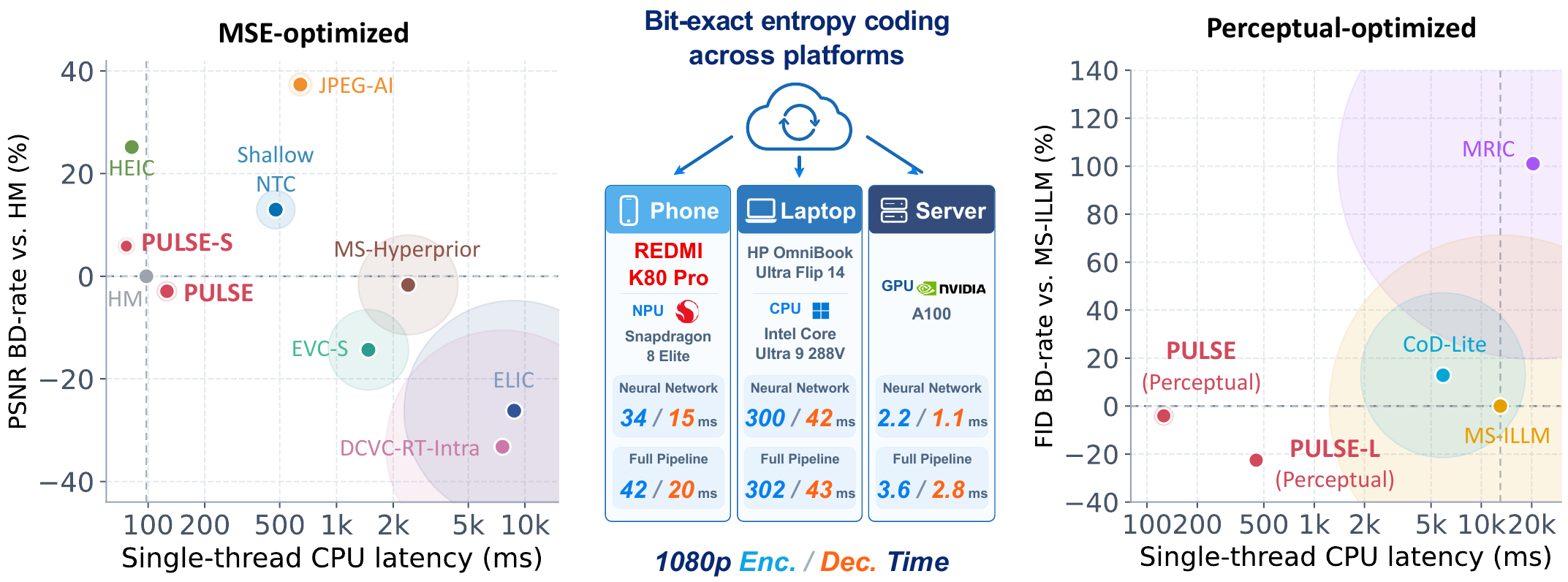}
    \caption{\textbf{\system targets practical image compression on resource-constrained hardware.}
    Its 5.2~kMAC/pixel receiver decodes a 1080p image in 126~ms on a single thread of an AMD EPYC 9V84, while enabling bit-exact entropy coding across devices like REDMI K80 Pro.
    On the CLIC Professional validation set, \system matches HM and its perceptual variant competes with larger codecs such as MS-ILLM~\citep{muckley2023illm}.
    \textit{Latency is measured for the full decoding pipeline, including network inference and entropy decoding. Circle radius denotes the decoding complexity.}}
    \label{fig:overview}
\end{figure*}

\section{Introduction}
\label{sec:intro}

Pioneering learned image compression systems~\citep{balle2017end,balle2018scale,minnen2018joint} used neural receivers requiring roughly 80~kMAC/pixel.
Subsequent work improved rate--distortion performance by scaling receiver complexity to several hundred kMAC/pixel~\citep{cheng2020learned,he2022elic,jiang2025mlicpp,liu2023tcm}, making real-time decoding challenging even on GPUs.
To bridge this deployment gap, instance-adaptive codecs reduce receiver complexity to below 5~kMAC/pixel, but shift substantial computation to per-image optimization during encoding~\citep{ladune2023coolchic,kim2024c3}.
Another line of work pursues efficient end-to-end codecs: EVC reduces receiver complexity to about 110~kMAC/pixel, Shallow NTC to approximately 20~kMAC/pixel, while DCVC-RT reorganizes computation to achieve real-time GPU throughput~\citep{wang2023evc,yang2023shallow,jia2025dcvcrt}.
Despite these advances, such systems remain challenging to deploy at scale on resource-constrained edge devices, particularly CPUs.

We propose \system to target the missing operating point: an ultra-low-complexity, highly parallel and end-to-end codec, with a sender and a receiver around 45 and 5.2 kMAC/pixel, respectively.
We use \textit{a single CPU thread} as a stringent deployment stress test, while retaining bounded-pass entropy dependencies and accelerator-friendly operators so that the same model can benefit from NPUs and GPUs for practical implementation.
We make \system practical by solving three core questions.

\textbf{How to design a codec under an ultra-low-complexity budget?}
Under such a stringent constraint, every channel and operator must be placed where it brings the greatest compression benefit.
PICO searches over millions of architectures, incurring substantial candidate-training costs and relying on statically predefined networks~\citep{tatwawadi2026pico}.
We instead introduce an \textit{agentic evolution} process that uses heuristic probes to inform architecture design through human--LLM collaboration.
With only seven evolution rounds and 22 training runs, \system achieves compression performance comparable to HM on the CLIC Professional validation set.
It decodes a 1080p image in 126~ms on a single thread of an AMD EPYC 9V84 CPU. 

\textbf{How to maintain bit-exact entropy coding across devices with minimal loss?}
Platform-dependent floating-point round-off can change the selected CDF and desynchronize the remaining bitstream.
\system addresses this with \textit{one integer linear projection} from the hyperlatent to the main-latent CDF bucket, which requires only 1.6~ms for a 1080p image on a single CPU thread.
In practice, this compact entropy-control path can execute on CPU for bit-exact entropy coding, while the remaining parts run efficiently on accelerators.
With a meta prior to improve hyperlatent coding, this scheme outperforms existing integerization-based methods in rate-distortion loss.

\textbf{How to effectively improve perceptual quality?}
We separate perceptual training into two stages: an LPIPS-augmented pre-training stage to establish detail synthesis capability, and a preference-optimization stage that jointly improves pixel fidelity, distribution alignment, and text and facial quality.
The optimized model competes with larger perceptual codecs.

Together, these components address key bottlenecks in practical image compression
on resource-constrained hardware. As illustrated in Figure~\ref{fig:overview},
\system enables practical compression across phones, laptops, and servers with low computational cost. 
On edge devices like REDMI K80 Pro, it takes 42 ms for encoding and 20 ms for decoding at 1080p. We expect further system-level optimization to improve its efficiency for real-world applications.

%% file: sec/2_architecture.tex
\section{PULSE Architecture}
\label{sec:architecture}

\begin{figure*}[t]
\centering
\includegraphics[width=\textwidth]{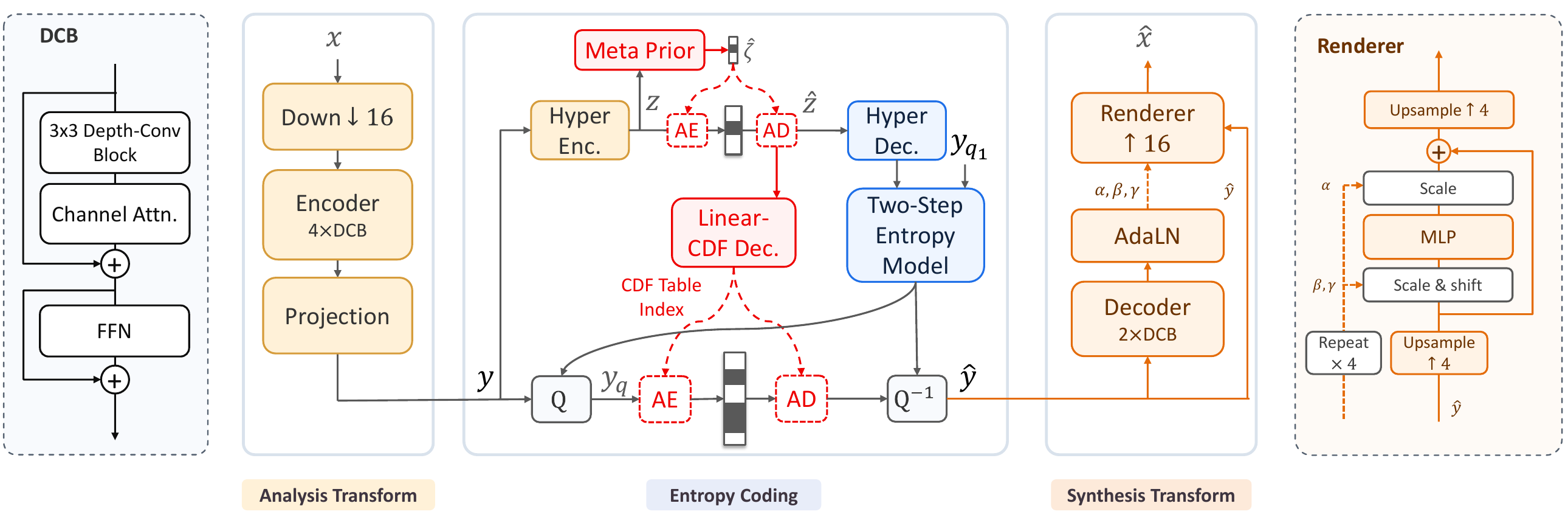}
\caption{\textbf{PULSE architecture.} It consists of an encoder at $1/16$ resolution, a decoupled linear CDF index decoder with a meta prior, a two-step entropy model, and a decoder with an AdaLN-modulated pixel renderer. AE and AD denote arithmetic encoding and decoding, respectively.}
\label{fig:architecture}
\end{figure*}

\system is an asymmetrical, variable-rate, cross-platform codec designed for ultra-low-complexity decoding. As illustrated in Figure~\ref{fig:architecture}, it consists of an analysis transform, a two-step entropy model, and a synthesis transform. All modules are designed with highly parallel computation to enable efficient implementation across different hardware platforms. We build it with depthwise convolution blocks (DCB) with a channel attention layer~\citep{hu2018senet}, which is used throughout modules.

\textbf{Analysis transform.}
Given an image $x\in[0,1]^{3\times H\times W}$, a stride-16 patch embedding produces a feature at $1/16$ spatial resolution. Four DCBs are then applied, followed by a $1\times1$ projection: 
\begin{equation}
y = \operatorname{Encoder}(x)
\in\mathbb R^{C_y\times H/16\times W/16}.
\end{equation}

\textbf{Entropy model and coding.}
Unlike the raster-ordered entropy model of Cool-Chic~\citep{ladune2023coolchic}, \system uses two bounded passes. A hyper encoder~\citep{balle2018scale} maps $y$ to a hyperlatent $z$ at $1/64$ input resolution. The rounded
$\hat z$ is coded by a QP-conditioned Meta Prior, which selects among a compact bank of factorized priors~\citep{balle2017end,jia2025dcvcrt}, and a hyper decoder restores a $C_y$-channel feature at the resolution of $y$:
\begin{equation}
z=\operatorname{HyperEncoder}(y)
\in\mathbb R^{C_z\times H/64\times W/64},\qquad
h=\operatorname{HyperDecoder}(\hat z).
\end{equation}

Following the dual spatial prior of DCVC-HEM~\citep{li2022hybrid}, a spatial--channel checkerboard partitions $y$ into complementary masks $m_0$ and $m_1$. The first entropy step predicts a quantization step $q$ and a mean $\mu_0$ from $h$. After decoding the first part, the second step refines $\mu_1$ from $(h,\hat y_0)$, then
\begin{equation}
y_{q_i}=\operatorname{round}\!\left(y/q-\mu_i\right)\odot m_i,
\qquad
\hat y_i=q\left(y_{q_i}+\mu_i\right)\odot m_i,
\qquad
\hat y=\hat y_0+\hat y_1.
\label{eq:two-part-quantization}
\end{equation}
The centered symbols are modeled by a discretized zero-mean Gaussian. Scale prediction is independent of the nonlinear mean path: one projection maps $\hat z$ directly to a continuous coordinate in a shared Gaussian CDF table.
Both passes reuse these scale coordinates, while only their $\mu_i$ differ.
Section~\ref{sec:cross-platform} specifies the Meta Prior and deterministic integer implementation.

% To further facilitate cross-platform implementation, \system decouples scale prediction from nonlinear mean prediction. A linear projection directly maps $\hat{z}$ to the index of a precomputed Gaussian CDF table, which is shared by the arithmetic encoder and decoder. More details are in Section~\ref{sec:cross-platform}.

\textbf{Synthesis transform.}
The synthesis transform separates spatial feature processing from patch-wise pixel rendering. The decoder consisting of two DCBs first processes $\hat{y}$ for spatial correlation modeling. In parallel, a content branch upsamples $\hat y$ by four to a $1/4$-resolution rendering grid. The low-resolution body generates AdaLN modulation parameters $(\alpha,\beta,\gamma)$~\citep{peebles2023dit} for a residual pointwise MLP on this grid. Finally, every grid vector is projected to a $4\times4$ RGB patch and rearranged by PixelShuffle~\citep{shi2016subpixel}:
\begin{equation}
\hat{x} =
\operatorname{Renderer}\!\left(
\hat{y},
\operatorname{Decoder}(\hat{y})
\right).
\end{equation}

%% file: sec/3_crossplatform.tex
\section{Bit-Exact Linear CDF Decoding with a Meta Prior}
\label{sec:cross-platform}

Platform-dependent floating-point round-off is particularly dangerous inside an entropy decoder: a single change in a CDF-table index can desynchronize every subsequent rANS symbol~\citep{duda2013ans}.
The conventional solution is to integerize the entire entropy model~\citep{he2022crossplatform}, ensuring deterministic execution of the CPU subgraph from $\hat z$ to CDF.
However, as shown in Table~\ref{tab:integer}, this introduces substantial CPU overhead and quantization loss under an ultra-low complexity budget.
In \system, we instead simplify the entropy-control path to the minimum computation required for bit-exact CDF selection while preserving coding efficiency.

\textbf{Linear scale decoder.}
We first follow~\cite{parnamaa2026mlvc} to derive the scale directly from $\hat z$ using a learned LUT, which is extremely fast but incurs a substantial rate-distortion loss.
We then integerize only the independent hyper scale decoder~\citep{ascenso2023jpegai}, rather than the entire entropy model, which effectively alleviates both issues.
This inspires us to take a step further: \textit{can we significantly simplify the hyper scale decoder while retaining its compression ratio?}
Surprisingly, we find that a simple \textit{linear projection} from $\hat z$ to the scale provides an effective middle ground.
The linear decoder is more robust to quantization, incurring only about 1.5\% BD-rate loss at INT8 precision compared to the full hyper scale decoder, while reducing CPU time from 28.1~ms to 1.2~ms.

\begin{figure*}[t]
    \centering
    \includegraphics[width=\textwidth]{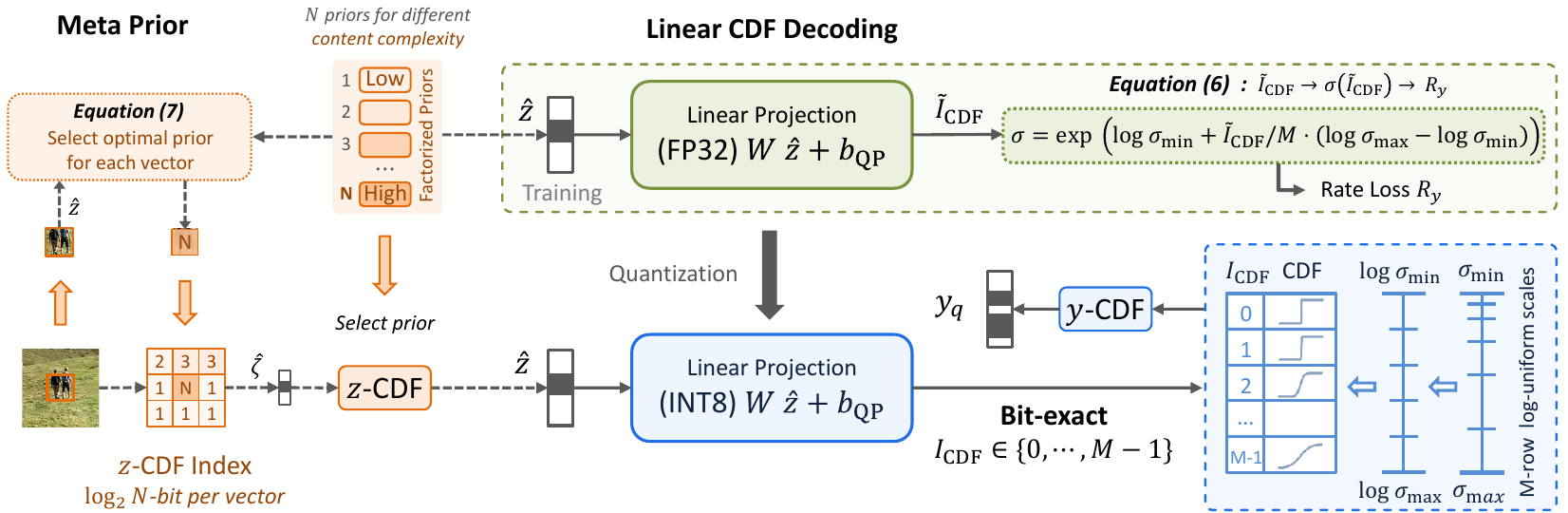}
    \caption{\textbf{Bit-exact entropy coding path.}
    Meta Prior transmits $\hat \zeta$ to select CDFs for $\hat z$ with different content complexities.
    An integer linear projection maps $\hat z$ to CDF indices for entropy coding of $y_q$.}
    \label{fig:linear_scale_decoder}
\end{figure*}

As shown in Figure~\ref{fig:linear_scale_decoder}, CDF selection based on $\log \sigma$ reveals two limitations.
Firstly, the $\sigma\to\log \sigma$ path must also execute in INT8, adding 2.9~ms of latency.
Secondly, uniform quantization makes $\hat{z}$ a fixed-step additive representation, whereas Gaussian CDF indices are uniformly spaced in log-scale. Consequently, affine scale decoding induces a scale-dependent CDF-index sensitivity, with the same latent step causing larger index changes at small scales and smaller changes at large scales.

\textbf{Linear CDF decoder.}
To address this, we propose a simple but effective solution: directly predict the CDF index instead of scale.
Specifically, a linear projection predicts a continuous coordinate $\tilde{I}_{CDF}$ in a shared M-row log-spaced Gaussian CDF table:
\begin{equation}
\tilde{I}_{CDF}=W\hat z+b_{\mathrm{QP}},\qquad
I_{\mathrm{CDF}}=\left\lfloor\operatorname{clip}(\tilde{I}_{CDF},0,M-1)\right\rfloor .
\label{eq:linear-cdf}
\end{equation}
Here $b_{\mathrm{QP}}$ is a learned per-QP bias.
During training, the bounded coordinate is mapped to
\begin{equation}
\sigma(\tilde{I}_{CDF})=\exp\left(
\log\sigma_{\min}
+\tilde{I}_{CDF}/M\cdot(\log\sigma_{\max}-\log\sigma_{\min})
\right)
\end{equation}
for differentiable Gaussian likelihood evaluation. We empirically set $\sigma_{\rm min}=0.11$, $\sigma_{\rm max}=256$ and $M=64$.
At inference, entropy coding consumes $I_{\mathrm{CDF}}$ directly, eliminating the runtime exponential and logarithm.
Linear CDF decoding eliminates the $\log \sigma$ computation to reduce CPU time to 1.6~ms, while achieving a lower BD-rate than the INT8 hyper scale decoder on the CLIC test set.

\textbf{Meta Prior.}
The extreme simplification of the linear CDF decoder shifts more rate-modeling responsibility onto $\hat z$.
However, the inflexible factorized prior uses the same prior distribution for $\hat z$ across all images and cannot adapt to different image contents.
Empirically, we observe that $\hat z$ accounts for about half of the bitstream for images with low content complexity.
As shown in Table~\ref{tab:integer}, it leads to a substantial degradation on the low-content-complexity (LC) subset.

To address this issue, we introduce a meta prior that enables content-adaptive entropy modeling with minimal additional decoding complexity.
Specifically, after training the entire model, we post-train only the factorized prior into a bank of $N$ prior models.
Each prior model is fitted to hyperlatent vectors from regions of different content complexity, with other modules fixed.
We then allow each $\hat z$ vector at a different spatial position to select its own prior model, using an expectation-maximization (EM) procedure to jointly optimize the prior models and their assignments. More details are provided in Appendix~\ref{app:training-details}.
At inference, for each spatial position $r$ in $\hat z$, the encoder evaluates the quantized-CDF coding cost $\ell_{q,k}(\hat z_r)$ under every prior bank and transmits the best bank index:
\begin{equation}
k_r^\star=\arg\min_{k\in\{0,\ldots,N-1\}}\ell_{q,k}(\hat z_r),\qquad
R_z=\sum_r \left[\log_2 N+\ell_{q,k_r^\star}(\hat z_r)\right].
\label{eq:meta-prior}
\end{equation}
A single $\log_2 N$-bit index is shared by all channels at each spatial position, serving as meta-prior information.
The decoder reads this metadata and selects the corresponding fixed CDF bank to decode $\hat z_r$ without additional neural-network inference.
Unlike some prior works that select competing priors for main latents~\citep{brummer2021end}, meta prior adapts $\hat z$ coding to compensate for the reduced capacity of the linear CDF decoder, particularly in low-content-complexity regions.
This brings a 9.8\% bitrate saving on low-complexity images and a 2.4\% saving on the overall set.

In deployment, the linear CDF activation and weights are INT8, accumulation is INT32, and the output is a UINT8 CDF index.
Meta Prior bank indices are coded with fixed-length coding, and can be further improved with merge coding.

\begin{table*}[t]
\centering
\caption{\textbf{Ablation of bit-exact entropy-coding designs.}
BD-rate is measured on the CLIC test set and a subset of images with low content complexity (LC subset).
Latency is measured for a 1080p image on one AMD EPYC 9V84 CPU thread.}
\label{tab:integer}
\small
\setlength{\tabcolsep}{3.8pt}
\begin{tabular}{l|cc|cc|cc}
\toprule
\multirow{2}{*}{Method} &
\multicolumn{2}{c|}{CLIC test set} &
\multicolumn{2}{c|}{LC subset} &
\multicolumn{2}{c}{Single-thread CPU Time} \\
& Float & INT8 & Float & INT8 & $\hat z\to \sigma$ & $\hat z\to$ CDF \\
\midrule
Integerize full entropy model
    & 1.5\% & 5.7\% & 3.1\% & 7.8\% & 85.9 ms & 86.0 ms \\
\textcolor{gray}{+ \sout{LUT scale decoder}}~\citep{parnamaa2026mlvc}
    & \textcolor{gray}{29.7\%} & \textcolor{gray}{29.7\%} & \textcolor{gray}{59.6\%} & \textcolor{gray}{59.6\%} & \textcolor{gray}{0.5 ms} & \textcolor{gray}{0.5 ms} \\
+ Hyper scale decoder~\citep{ascenso2023jpegai}
    & \textbf{--0.1\%} & 3.2\% & 2.1\% & 6.4\% & 28.1 ms & 29.2 ms \\
\midrule
+ Linear scale decoder
    & 4.5\% & 4.5\% & 11.8\% & 12.0\% & 1.2 ms & 4.1 ms \\
+ Linear CDF decoder
    & 2.4\% & 2.4\% & 9.8\% & 9.9\% & N/A & 1.6 ms \\
+ Meta Prior $\to$ \textbf{Proposal}
    & 0.0\% & \textbf{0.04\%} & \textbf{0.0\%} & \textbf{0.04\%} & N/A & 1.6 ms \\
\bottomrule
\end{tabular}
\end{table*}

%% file: sec/4_agentic_evolution.tex
\section{Agentic Evolution with Heuristic Probes}
\label{sec:agentic-evolution}

\begin{figure*}[t]
    \centering
    \includegraphics[width=\textwidth]{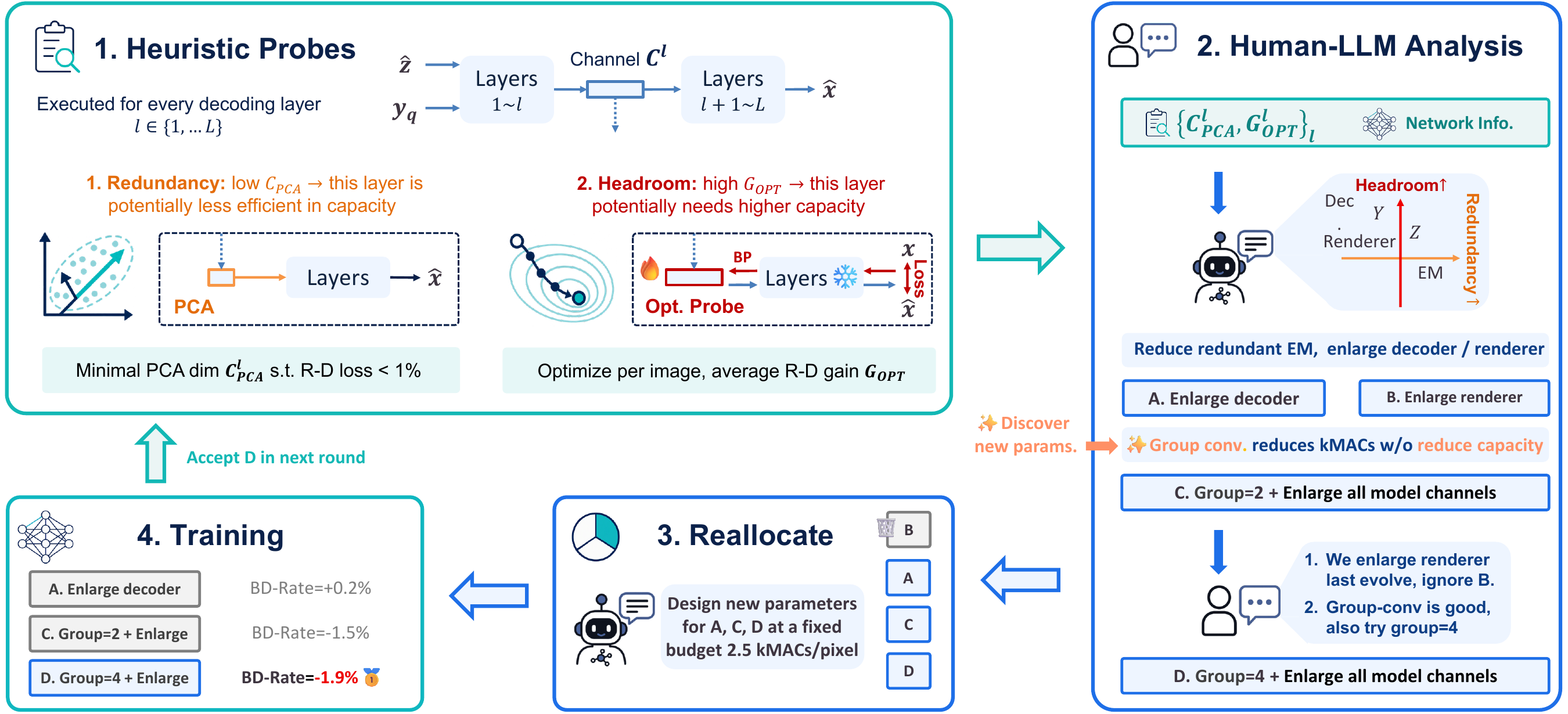}
    \caption{\textbf{Agentic evolution}, illustrated for round E4$\rightarrow$E5 in the complete trajectory in Figure~\ref{fig:agentic_evolution_res}.
    Layer-wise probes measure \textcolor[HTML]{E46C0A}{redundancy} and \textcolor[HTML]{C00000}{headroom}.
    A human researcher and an LLM agent interpret the evidence to propose fixed-budget reallocations, and validate them by training.}
    \label{fig:agentic_evolution_method}
\end{figure*}

Under an ultra-low complexity budget, every channel and operator matters, making architecture-parameter allocation critical to compression performance.
We evolve the neural receiver through four stages: \textit{probe}, \textit{analyze}, \textit{reallocate}, and \textit{train}, as illustrated in Figure~\ref{fig:agentic_evolution_method}.
The analysis transform is fixed throughout the process to isolate receiver design.

\subsection{Heuristic Representation Probes}

Given a trained network, we probe representations at selected interfaces. For layer $l\in\{1,\cdots,L\}$, we examine the input latent $a^l$ with $C^l$ channels. We measure two complementary properties.

\textbf{Redundancy probe with PCA dimension.}
The latent $a^l$ may contain redundant channels if its allocated dimension exceeds what is required to preserve end-task performance.
We fit a dataset-shared PCA basis and reconstruct it from its first $c$ principal components, denoted by $\Pi_c(a^l)$.
We then find the smallest dimension whose relative rate--distortion loss increase is at most $\epsilon=1\%$:
\begin{equation}
C_{\mathrm{PCA}}^l =
\min_c\left\{
c:
\mathcal{L}_{\mathrm{RD}}(\Pi_c(a^l))
-\mathcal{L}_{\mathrm{RD}}(a^l)
\le \epsilon\cdot\mathcal{L}_{\mathrm{RD}}(a^l)
\right\}.
\label{eq:probe-pca}
\end{equation}
The ratio $C_{\mathrm{PCA}}^l/C^l$ suggests utilization of the allocated channel subspace.
A small ratio indicates potentially reclaimable channel redundancy.

\textbf{Headroom probe with latent optimization.}
The PCA probe suggests redundancy but does not assess whether upstream computation produces the best representation the downstream network can exploit.
We therefore optimize $a^l$ independently for each image while freezing all downstream parameters.
For decoder-side interfaces, the coded symbols are already fixed, so this intervention keeps the rate unchanged and directly measures an oracle PSNR gain:
\begin{equation}
a^{l*}=\arg\min_{a^l}
D\!\left(F_{>l}(a^l),x\right),
\qquad
G_{\mathrm{OPT}}^l =
\operatorname{PSNR}\!\left(F_{>l}(a^{l*}),x\right)
-\operatorname{PSNR}\!\left(F_{>l}(a^l),x\right).
\label{eq:probe-headroom}
\end{equation}
Entropy-side interfaces use the rate--distortion objective, as detailed in Appendix~\ref{app:agentic-details}.
A large $G_{\mathrm{OPT}}^l$ indicates that the downstream network can benefit from a more informative value at this interface.

\subsection{Human--LLM Collaborative Analysis}

The LLM agent receives ${\{C_\mathrm{PCA}}^l,G_{\mathrm{OPT}}^l\}_{l=1}^{L}$ together with the network graph, parameter sharing, exact component costs, and the history of previous rounds.
With this comprehensive context, the LLM interprets the probe results in the architectural context and proposes fixed-budget changes for empirical validation.
The human researcher then reviews the hypotheses, implementation feasibility and budget accounting, and selects candidates for training.

In practice, some bottlenecks cannot be addressed by changing parameters in the predefined search space.
Both the LLM agent and the human researcher may therefore introduce a \textcolor[HTML]{FF9065}{spark}: a new architectural mechanism or adjustable factor motivated by the probe evidence. This allows the
optimization space itself to evolve as new bottlenecks are discovered.

\textbf{\textcolor{blue}{Discussion}: comparison with NAS.} Agentic evolution differs from conventional NAS in two respects.
First, heuristic probes help prioritize candidate directions before from-scratch training instead of enumerating a large static search space.
Second, the human--LLM process is open-ended: probe evidence can motivate new parameters or mechanisms through sparks rather than restricting all candidates to search axes fixed in advance.

\subsection{Fixed-budget Reallocation and From-Scratch Training}

Each selected proposal is profiled before training to ensure that parameter reallocation remains within a fixed complexity budget.
The remaining candidates are trained from scratch with the same data, rate range, optimizer, and schedule.
Finally, the candidate with the best compression performance is selected as the promotion for the next round.

\subsection{Evolution at 2.5 kMAC/pixel and Scaling to Different Sizes}

The complete trajectory of seven generations is reported in Appendix~\ref{app:agentic-details} (Figure~\ref{fig:agentic_evolution_res}).
Across the full trajectory, human--LLM collaboration alternates between reallocating known parameters and introducing new search axes. 
In detail, the depth and width of different modules are jointly adjusted step by step, and three rounds discover new parameters including the renderer resolution, group convolution and FFN width. 
The final E7 achieves a 10.2\% BD-rate reduction relative to E1.
Finally, we adjust the channel dimensions to multiples of 16 for efficient inference of \system-S, and then enlarge channels to scale the decoding complexity to 5.2 and 20~kMAC/pixel, resulting in the \system and \system-L models, respectively.

\textbf{\textcolor{blue}{Discussion}: scope and generalization.}
We study human--LLM co-design for ultra-low-complexity image codecs, rather than task-agnostic architecture search.
The workflow is history-dependent, with insights from earlier rounds guiding subsequent decisions.
Thus, the contributions of human and LLM agents cannot be cleanly separated post hoc.
Within this setting, the workflow enables effective architecture evolution with few training runs.
Generalization to other tasks may require task-specific search spaces and objectives, which we leave for future work.

%% file: sec/5_perceptual.tex
\section{Perceptual and Region-of-Interest Optimization}
\label{sec:perceptual}

This section introduces the techniques used to improve the subjective
reconstruction quality of \system. When optimized with MSE alone, the decoder
tends to over-smooth uncertain details, resulting in blurry textures and local
structures. Inspired by perceptual codecs~\citep{mentzer2020hific,muckley2023illm},
we adopt a two-stage perceptual optimization process.

\textbf{Stage I: LPIPS-First Perceptual Pre-Training.} The first stage establishes the model's detail synthesis capability by emphasizing the LPIPS loss~\citep{zhang2018lpips}:
\begin{equation}
\mathcal L_{\rm Stage\ I}=R_y+R_z+\lambda
\left[
\phi_{\rm MSE} D_{\rm MSE}
+\phi_{\rm LPIPS}^{\rm I} D_{\rm LPIPS}
\right].
\label{eq:perceptual-pretrain}
\end{equation}
where $\phi_{\rm LPIPS}^{\rm I}$ is set substantially higher than in prior work~\citep{muckley2023illm} to encourage stronger perceptual detail synthesis.

\textbf{Stage II: Preference Optimization.} The second stage aligns reconstruction quality with perceptual preference criteria.
We make three modifications: (1) reduce the LPIPS weight back to mitigate its artifacts and imperfect perceptual alignment; (2) introduce adversarial training with a DINOv2-based discriminator~\citep{oquab2023dinov2} for distribution-level alignment; and (3) introduce region-of-interest (ROI) optimization for semantically important regions, e.g., small text and face regions.

We pre-detect text- and face-rich images and store their corresponding ROI
boxes. Within these regions, we increase the MSE and LPIPS weights and introduce
semantic feature alignment:
\begin{equation}
\begin{split}
\mathcal L_{\rm Stage\ II}=R_y+R_z+\lambda\big[
\phi_{\rm MSE} D_{\rm MSE}
+\phi_{\rm LPIPS}^{\rm II} D_{\rm LPIPS}
+\phi_{\rm GAN} D_{\rm GAN}
+\phi_{\rm ROI} D_{\rm ROI}
\big].
\end{split}
\label{eq:preference-objective}
\end{equation}
\begin{equation}
D_{\rm ROI}=\phi_{\rm FaceNet} D_{\rm FaceNet}+\phi_{\rm CRNN} D_{\rm CRNN},
\label{eq:roi-feature-weak}
\end{equation}

where $\phi_{\rm LPIPS}^{\rm I}=4\times\phi_{\rm LPIPS}^{\rm II}$.
FaceNet~\citep{schroff2015facenet} uses square face crops with contextual
regions, while the CRNN-CNN teacher~\citep{shi2017crnn} uses
aspect-preserving text crops.
This stage jointly improves pixel fidelity, perceptual quality, and ROI quality, which is verified in Table~\ref{tab:percetual_ablation}.

%% file: sec/6_experiments.tex
\section{Experiments}
\label{sec:experiments}

\begin{figure*}[t]
    \centering
    \includegraphics[width=\textwidth]{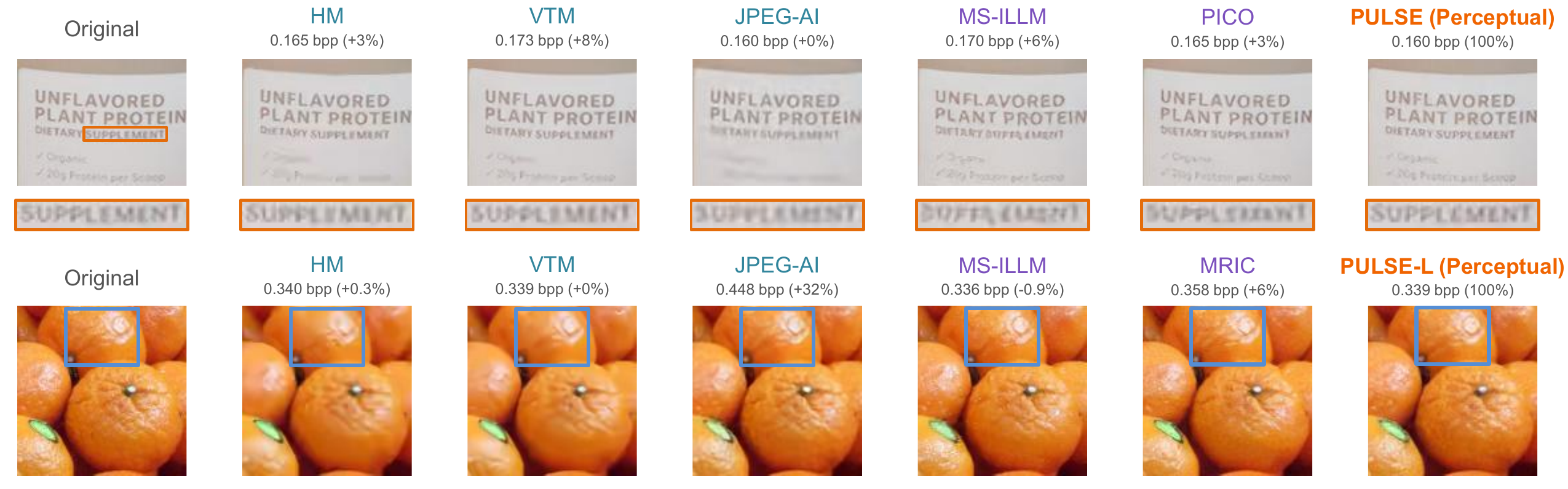}
    \caption{\textbf{Visual comparison.}
    \system achieves comparable or better visual quality after perceptual and ROI optimization. More results are in Appendix~\ref{app:visual-comparisons}.
    }
    \label{fig:visual_main}
\end{figure*}

\subsection{Experimental Setup}

\textbf{Training.}
\system is optimized using SOAP~\citep{vyas2025soap}, requiring 15 hours for the MSE version and 36 hours for the perceptual version on a single NVIDIA H100 GPU.
More details of training schedules and datasets are in Appendix~\ref{app:training-details}.
\system supports eight quantization parameters in a single model through the
module-bank formulation~\citep{jia2025dcvcrt}, and each quantization parameter is
trained with its corresponding $\lambda$.

\begin{figure*}[t]
    \centering
    \includegraphics[width=\textwidth]{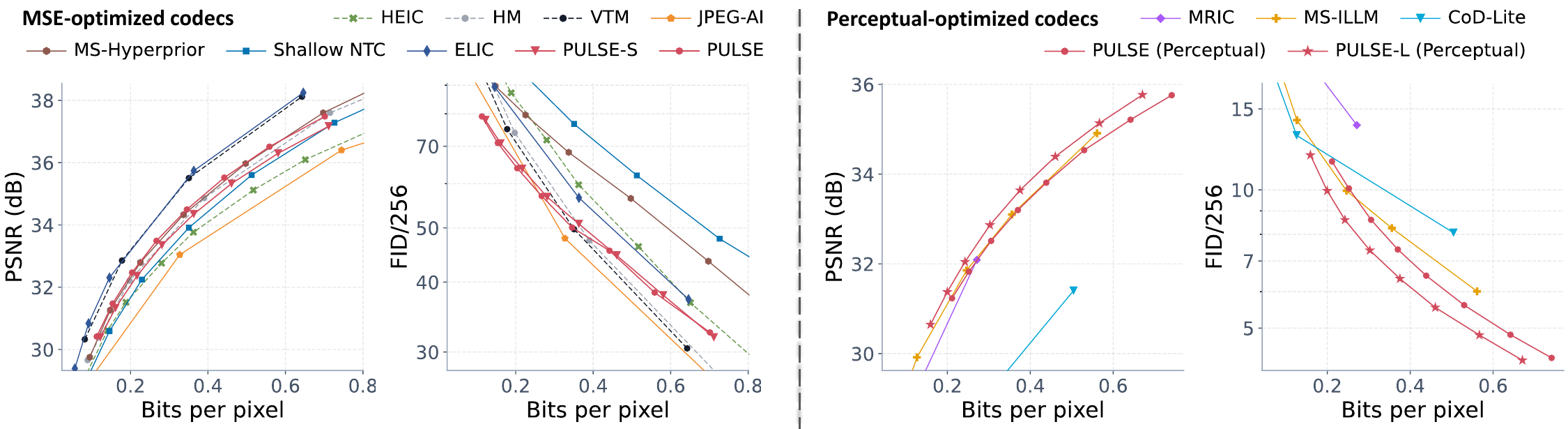}
    \caption{\textbf{Rate--distortion curves.} CLIC Professional validation results for MSE-optimized codecs (left) and
    perceptually optimized codecs (right). More results are in Appendix~\ref{app:rd-comparisons}.}
    \label{fig:rd_curves}
\end{figure*}

\definecolor{mseblue}{RGB}{70,125,200}
\definecolor{percorange}{RGB}{215,140,65}

\begin{table}[t]
\centering
\caption{\textbf{Ablation studies.} BD-Rate ($\downarrow$) is measured in terms of \textcolor{mseblue}{PSNR} and \textcolor{percorange}{FID} for the \textcolor{mseblue}{MSE-} and \textcolor{percorange}{perceptual-}optimized versions, respectively. Decoder speed is measured on 1080p images on an NPU.}
\label{tab:ablation}
\setlength{\tabcolsep}{5pt}

\begin{minipage}[t]{0.51\textwidth}
\begin{subtable}[t]{\linewidth}
    \centering
    \label{tab:ablation-structure}
    \small
    \begin{tabular}{@{}lccc@{}}
    \toprule
    Decoder structure & \textcolor{mseblue}{PSNR} & \textcolor{percorange}{FID} & Speed\\
    \midrule
    Progressive Upsampling & 1.1\% & 5.8\% & 0.9$\times$\\
    Single Low-Res. Decoder & 12.0\% & 18.4\% & 1.6$\times$\\
    \textbf{Low-Res. Decoder + Renderer} & 0.0\% & 0.0\% & 1.0$\times$\\
    \bottomrule
    \end{tabular}
\end{subtable}
\end{minipage}%
\begin{minipage}[t]{0.52\textwidth}
\begin{subtable}[t]{\linewidth}
    \centering
    \label{tab:ablation-search}
    \small
    \begin{tabular}{@{}lcc@{}}
    \toprule
    Architecture search
    & \makecell{\#Trains $\downarrow$}
    & \makecell{\textcolor{mseblue}{PSNR}} \\
    \midrule
    No Search & 1 & 0.0\% \\
    Hardware-aware NAS & 1280 & -7.5\% \\
    \textbf{Agentic Evolution} & 22 & -10.2\% \\
    \bottomrule
    \end{tabular}
\end{subtable}
\end{minipage}

\end{table}

\textbf{Benchmarks and metrics.}
We use the CLIC Professional validation set~\citep{clic2020} for comparison.
For MSE-optimized compression, we compare with Mean-Scale Hyperprior~\citep{minnen2018joint},
ELIC~\citep{he2022elic}, EVC~\citep{wang2023evc}, DCVC-RT-Intra~\citep{jia2025dcvcrt}, Shallow NTC~\citep{yang2023shallow}, HM-16.25~\citep{sullivan2012hevc}, and VTM-17.0~\citep{bross2021vvc}.
For perceptual compression, we compare with MRIC~\citep{agustsson2023mric}, MS-ILLM~\citep{muckley2023illm}, and CoD-Lite~\citep{jia2026codlite}. HiFiC and PICO comparisons on matched datasets are in the appendix.
We report PSNR and patch-based FID~\citep{heusel2017fid} ($256\times256$, overlapping)~\citep{mentzer2020hific}.
BD-rate follows \citet{bjontegaard2001}.
We evaluate $\mathrm{bpp}\ge0.05$, so ultra-low-bitrate results may differ from prior reports.

\textbf{Complexity.}
We report the parameters and multiply--accumulate operations exercised
by each sender and receiver.
Latency is measured at $1088\times1920$ on a single AMD EPYC 9V84 CPU thread.

\subsection{Experimental Results}

\textbf{Comparison results.} Figure~\ref{fig:rd_curves} and Table~\ref{tab:complexity-latency} report comparisons with MSE- and perceptual-optimized codecs.
\system-S encodes a 1080p image in 920~ms and decodes it in 77~ms, faster than all compared codecs, including software-implemented HEIC. It also achieves a lower PSNR-based BD-rate of $+5.9\%$, compared to $+25.2\%$ for HEIC.
\system achieves a $-2.9\%$ BD-rate relative to HM-16.25 with $9\times$ faster encoding, while competing with the much larger MS-ILLM after perceptual optimization.
\system-L achieves a $-22.5\%$ FID-based BD-rate relative to MS-ILLM with substantially fewer computations.
Figure~\ref{fig:visual_main} shows text and natural-texture examples for visual comparison, and more examples are in the appendix.

\textbf{Ablation study.} As shown in Table~\ref{tab:ablation}, under the same decoding complexity, our rendering-based decoder outperforms the progressive upsampling decoder~\citep{balle2017end} in both compression efficiency and NPU inference speed.
The single low-resolution decoder~\citep{jia2025dcvcrt} is faster but incurs a relatively large performance loss.
Our scheme achieves the best overall trade-off.
Compared to hardware-aware NAS~\citep{tatwawadi2026pico}, the proposed agentic evolution uses fewer training runs to achieve better performance.

\newlength{\complexitygroupsep}
\setlength{\complexitygroupsep}{10.5pt}
\newcolumntype{G}{@{\hspace{\complexitygroupsep}}}
\newcommand{\metriccell}[2]{%
  {\setlength{\fboxsep}{1.2pt}%
  \colorbox{#1}{\makebox[4.2em][c]{\strut #2}}}%
}
\newcommand{\msecell}[1]{\metriccell{oursblue!7}{#1}}
\newcommand{\perccell}[1]{\metriccell{oursorange!8}{#1}}

\begin{table*}[t]
\centering
\caption{\textbf{Complexity and rate--distortion performance.} Time measurements use $1088\times1920$ inputs on a single-thread CPU. BD-rate is evaluated on the CLIC Professional validation set. VBR denotes variable bitrates within one model, and BE denotes bit-exact entropy coding across platforms.}
\label{tab:complexity-latency}
\scriptsize
\setlength{\tabcolsep}{2.7pt}
\resizebox{\textwidth}{!}{%
\begin{tabular}{lcccGcccGcccGcc}
\toprule
\multirow{2}{*}{Method} & \multirow{2}{*}{VBR} & \multirow{2}{*}{BE} & \multirow{2}{*}{Target} & \multicolumn{3}{c}{Encoding} & \multicolumn{3}{c}{Decoding} & \multicolumn{2}{c}{BD-rate $\downarrow$} \\
\cmidrule(lr){5-7}\cmidrule(lr){8-10}\cmidrule(lr){11-12}
& & & & Params & kMAC/px & Time & Params & kMAC/px & Time & \makecell{on PSNR} & \makecell{on FID} \\
\midrule
HEIC & $\checkmark$ & $\checkmark$ & \msecell{MSE} & \multicolumn{2}{c}{\textcolor{gray}{libheif + x265}} & 1.4 s & \multicolumn{2}{c}{\textcolor{gray}{libheif + libde265}} & 80 ms & \msecell{+25.2\%} & \textcolor{gray}{+2680\%} \\
HM-16.25 & $\checkmark$ & $\checkmark$ & \msecell{MSE} & \multicolumn{2}{c}{\textcolor{gray}{HM Encoder}} & 8.8 s & \multicolumn{2}{c}{\textcolor{gray}{HM Decoder}} & 98 ms & \msecell{0.0\%} & \textcolor{gray}{+2450\%} \\
VTM-17.0 & $\checkmark$ & $\checkmark$ & \msecell{MSE} & \multicolumn{2}{c}{\textcolor{gray}{VTM Encoder}} & 249 s & \multicolumn{2}{c}{\textcolor{gray}{VTM Decoder}} & 189 ms & \msecell{-23.4\%} & \textcolor{gray}{+2660\%} \\
\midrule
JPEG-AI & \textcolor{gray}{$\times$} & $\checkmark$ & \msecell{MSE} & 7.1 M & 56 & 2.6 s & 4.4 M & 8.1 & 643 ms & \msecell{+37.4\%} & \textcolor{gray}{+1720\%} \\
MS-Hyperprior & \textcolor{gray}{$\times$} & \textcolor{gray}{$\times$} & \msecell{MSE} & 14 M & 112 & 2.1 s & 12 M & 109 & 2.4 s & \msecell{-1.7\%} & \textcolor{gray}{+3830\%} \\
ELIC & \textcolor{gray}{$\times$} & \textcolor{gray}{$\times$} & \msecell{MSE} & 26 M & 335 & 6.5 s & 24 M & 332 & 8.8 s & \msecell{-26.2\%} & \textcolor{gray}{+3430\%} \\
EVC-S & $\checkmark$ & \textcolor{gray}{$\times$} & \msecell{MSE} & 12 M & 71 & 1.4 s & 11 M & 79 & 1.4 s & \msecell{-14.3\%} & \textcolor{gray}{+3040\%} \\
DCVC-RT-Intra & $\checkmark$ & \textcolor{gray}{$\times$} & \msecell{MSE} & 31 M & 258 & 5.4 s & 36 M & 362 & 7.6 s & \msecell{-33.2\%} & \textcolor{gray}{+3200\%} \\
Shallow NTC & \textcolor{gray}{$\times$} & \textcolor{gray}{$\times$} & \msecell{MSE} & 23 M & 277 & 5.6 s & 10 M & 21 & 476 ms & \msecell{+13.0\%} & \textcolor{gray}{+5590\%} \\
% N-O Cool-Chic & \textcolor{gray}{$\times$} & \textcolor{gray}{$\times$} & \msecell{MSE} & 1.0 M & 161 & 10 s & 0.0019 M & 2.3 & 1.3 s & -- & -- \\
\midrule
% HiFiC & \textcolor{gray}{$\times$} & \textcolor{gray}{$\times$} & \perccell{Perc.} & 25 M & 98 & 2.5 s & 140 M & 546 & 11 s & \textcolor{gray}{--} & \perccell{--} \\
MRIC & \textcolor{gray}{$\times$} & \textcolor{gray}{$\times$} & \perccell{Perc.} & 74 M & 823 & 16 s & 71 M & 829 & 20 s & \textcolor{gray}{+46.1\%} & \perccell{+101.0\%} \\
MS-ILLM & \textcolor{gray}{$\times$} & \textcolor{gray}{$\times$} & \perccell{Perc.} & 25 M & 98 & 2.2 s & 170 M & 675 & 13 s & \textcolor{gray}{+38.0\%} & \perccell{0.0\%} \\
CoD-Lite & \textcolor{gray}{$\times$} & $\checkmark$ & \perccell{Perc.} & 28 M & 344 & 7.0 s & 52 M & 219 & 5.9 s & \textcolor{gray}{+218.8\%} & \perccell{+12.9\%} \\
PICO & $\checkmark$ & $\checkmark$ & \perccell{Perc.} & 5.6 M & 75 & 2.0 s & 4.3 M & 46 & 1.2 s & \textcolor{gray}{N/A} & \perccell{N/A$^{\ddagger}$} \\
\midrule
\system-S & $\checkmark$ & $\checkmark$ & \msecell{MSE} & 13 M & 45 & 920~ms & 1.0 M & 2.7 & 77 ms & \msecell{+5.9\%} & \textcolor{gray}{+2530\%} \\
\multirow{2}{*}{\system} & \multirow{2}{*}{$\checkmark$} & \multirow{2}{*}{$\checkmark$} & \msecell{MSE} & 13 M & 45 & 920 ms & 1.7 M & 5.2 & 126 ms & \msecell{-2.9\%} & \textcolor{gray}{+2420\%} \\
& & & \perccell{Perc.} & 13 M & 45 & 920 ms & 1.7 M & 5.2 & 126 ms & \textcolor{gray}{+45.5\%} & \perccell{-4.1\%} \\
\system-L & $\checkmark$ & $\checkmark$ & \perccell{Perc.} & 15 M & 48 & 972~ms & 5.7 M & 20 & 420 ms & \textcolor{gray}{+30.8\%} & \perccell{-22.5\%} \\
\bottomrule
\end{tabular}
}
\par\vspace{2pt}
\begin{minipage}{\textwidth}
\scriptsize
$^{\ddagger}$ PICO complexity and timing use our local reproduction. We compare with its released 300-image reconstructions in Appendix~\ref{app:pico}.
\end{minipage}
\end{table*}

\subsection{Practical Implementation}

In prior sections, we focus on single-thread CPU execution as a pressure test.
To assess practical deployment, we evaluate \system across diverse platforms, including an Intel Core Ultra 9 288V CPU, NVIDIA A100 and H100 GPUs, and a Qualcomm Snapdragon 8 Elite NPU.
Executing the linear CDF decoding on the CPU~\textbf{verifies bit-exact entropy decoding} across all devices.
Figure~\ref{fig:overview} shows that the low complexity translates to high throughput on capable hardware.
On a REDMI K80 Pro with a Snapdragon NPU, the complete encoding and decoding pipelines take 42 ms and 20 ms, respectively, for 1080p images.
Overall, these results provide an indication that \system's ultra-low-complexity design can translate into practical deployment.

%% file: sec/7_landscape.tex
\section{Related Work}
\label{sec:related}

\textbf{Efficient neural codecs.}
Recent work reduces neural decoding complexity through lightweight contexts,
networks, and architectures. ELIC reduces serial context~\citep{he2022elic}, EVC uses mask decay~\citep{wang2023evc}, FastNIC targets low-complexity compression~\citep{zhang2025learning}, and DCVC-RT and Shallow NTC reduce operation complexity or depth~\citep{jia2025dcvcrt,yang2023shallow}.
We instead explicitly target ultra-low-complexity, end-to-end decoding on a single CPU thread.
Cool-Chic and C3 use tiny image-specific decoders, which require substantial per-image optimization for encoding~\citep{ladune2023coolchic,kim2024c3}. In addition, their highly sequential entropy model limits implementation efficiency.

\textbf{Perceptual neural codecs.}
Another line targets perceptual quality by modeling richer image details.
HiFiC and MS-ILLM enlarge synthesis networks to model realistic texture
\citep{mentzer2020hific,muckley2023illm}.
CoD-Lite enables real-time diffusion-based decoding, while PICO uses hardware-aware NAS to balance perceptual quality and efficiency~\citep{jia2026codlite,tatwawadi2026pico}.
However, perceptual compression on resource-constrained hardware remains underexplored.

\textbf{Cross-platform Compression.}
Post-training quantization can make probability prediction deterministic across
platforms~\citep{he2022crossplatform}. JPEG~AI standardizes integer hyper scale decoder~\citep{ascenso2023jpegai,tatwawadi2026pico}.
However, their scale decoding still relies on a large network. On devices with only an NPU as an accelerator, this bit-exact computation may fall back to the host CPU, where running such a large network is less efficient.
MLVC encodes scales into the hyperlatent using a lookup table, but incurs a considerable coding cost \citep{parnamaa2026mlvc}.
Thus, a simple-to-deploy scheme with low coding loss is still needed.

%% file: sec/8_conclusion.tex
\section{Conclusion}
\label{sec:conclusion}

\system demonstrates that practical learned image compression is possible under an extreme computational budget.
With a receiver complexity of 5.2~kMAC/pixel, it decodes a 1080p image in 126~ms on a single CPU thread while achieving HM-comparable compression performance.
Agentic evolution recovers performance under this tight budget, while a meta prior and integer linear CDF predictor enable bit-exact entropy coding across platforms.
Perceptual optimization further enables it to compete with larger perceptual codecs such as MS-ILLM.

\textbf{Limitation.}
Although we deploy \system across heterogeneous devices to demonstrate its practical applicability, further optimization is needed for real-world deployment.
While ROI processing improves visual quality for practical applications, \system is not specifically optimized for screen content, games, or artistic images, and its performance may degrade on these contents.
Thus, \system should be viewed as an academic prototype rather than a ready-to-use software system.

%% file: sec/X_supp.tex
\clearpage

\section{Model details}
\label{app:model-details}

This section introduces more detail of implementation of \system.

\subsection{Component Complexity Analysis}

Table~\ref{tab:component-full} reports the component accounting of \system.
The MSE and perceptual models use the same deployed topology and therefore have the same parameter and MAC counts.

\begin{table}[h]
\centering
\caption{Component accounting for \system.}
\label{tab:component-full}
\small
\setlength{\tabcolsep}{5pt}
\resizebox{\textwidth}{!}{%
\begin{tabular}{lcccccccc}
\toprule
\multirow{2}{*}{Component} & \multirow{2}{*}{Sender} & \multirow{2}{*}{Receiver} & \multicolumn{2}{c}{PULSE} & \multicolumn{2}{c}{PULSE-S} & \multicolumn{2}{c}{PULSE-L}\\
\cmidrule(lr){4-5}\cmidrule(lr){6-7}\cmidrule(lr){8-9}
~ & ~ & ~ & kMAC/px & Param. & kMAC/px & Param. & kMAC/px & Param.\\
\midrule
Analysis transform & $\checkmark$ & $\times$ & 43.22 & 12.1M & 43.08 & 11.1M & 43.21 & 12.1M\\
Hyper encoder & $\checkmark$ & $\times$ & 0.45 & 0.30M & 0.31 & 0.19M & 0.93 & 0.66M\\
Hyper decoder & $\checkmark$ & $\checkmark$ & 0.24 & 0.15M & 0.16 & 0.09M & 0.44 & 0.32M\\
Entropy model step-1 & $\checkmark$ & $\checkmark$ & 0.58 & 0.17M & 0.48 & 0.13M & 1.35 & 0.41M\\
Entropy model step-2 & $\checkmark$ & $\checkmark$ & 0.74 & 0.21M & 0.61 & 0.16M & 1.67 & 0.50M\\
Linear-CDF predictor & $\checkmark$ & $\checkmark$ & 0.12 & 0.50M & 0.08 & 0.33M & 0.18 & 0.74M\\
Synthesis transform & $\times$ & $\checkmark$ & 3.53 & 0.71M & 1.43 & 0.26M & 16.34 & 3.70M\\
\midrule
Neural sender & $\checkmark$ & $\times$ & 45.35 & 13.5M & 44.73 & 12.0M & 47.78 & 14.8M\\
Neural receiver & $\times$ & $\checkmark$ & 5.21 & 1.74M & 2.77 & 0.98M & 19.98 & 5.69M\\
\bottomrule
\end{tabular}%
}
\end{table}

\subsection{Latency Analysis}

Table~\ref{tab:cpu-full-pipeline} breaks down the end-to-end CPU latency of PULSE. Neural inference dominates the runtime, while non-neural processing takes only 7 ms for either encoding or decoding.
PULSE-S retains the same encoding latency but lowers end-to-end decoding latency from 126 ms to 77 ms, a 38.9\% reduction.
PULSE-L is slower due to increased complexity.

\begin{table}[h]
\centering
\caption{Single-thread CPU end-to-end latency of PULSE. \textit{Others} includes non-neural processing, such as entropy coding and CDF indexing.}
\label{tab:cpu-full-pipeline}
\small
\setlength{\tabcolsep}{4pt}
\begin{tabular}{llccc}
\toprule
Model & Pipeline & Neural inference & Others & Total \\
\midrule
\multirow{2}{*}{\system-s} & Encoding & 913 ms & 7 ms & 920 ms \\
& Decoding & 70 ms & 7 ms & 77 ms \\
\midrule
\multirow{2}{*}{\system} & Encoding & 913 ms & 7 ms & 920 ms \\
& Decoding & 119 ms & 7 ms & 126 ms \\
\midrule
\multirow{2}{*}{\system-L} & Encoding & 963 ms & 9 ms & 972 ms \\
& Decoding & 411 ms & 9 ms & 420 ms \\
\bottomrule
\end{tabular}
\end{table}

\subsection{Cross-device Entropy Coding validation}
\label{app:cross-platform-validation}

As shown in Figure~\ref{fig:overview}, we validate the linear CDF decoded index and latent-symbol decoding paths on three platforms:
(1) a REDMI K80 Pro phone using the Snapdragon 8 Elite NPU,
(2) an HP OmniBook Ultra Flip 14 laptop using only its Intel Core Ultra 9 288V CPU, and
(3) a server using an NVIDIA A100 GPU.
Although the laptop also includes an NPU, we disable NPU acceleration to characterize CPU-only performance.
For all devices, we deploy entropy coding and linear CDF decoding on there CPUs.
Table~\ref{tab:cross-device-validation} reports zero CDF-index and decoded-symbol mismatches.

\begin{table}[h]
\centering
\caption{Cross-device entropy coding bit-exact validation. Mismatch counts both CDF-index and decoded-symbol mismatches.}
\label{tab:cross-device-validation}
\small
\setlength{\tabcolsep}{5pt}
\resizebox{\textwidth}{!}{%
\begin{tabular}{llllc}
\toprule
Environment & Platform & Neural backend & CPU backend & Mismatch \\
\midrule
Phone & REDMI K80 Pro & NPU: Snapdragon 8 Elite & Snapdragon 8 Elite & \textbf{0} \\
Laptop & HP OmniBook Ultra Flip 14 & CPU: Intel Core Ultra 9 288V & Intel Core Ultra 9 288V & \textbf{0} \\
Server & GPU server & GPU: NVIDIA A100 & AMD EPYC 9V84 & \textbf{0} \\
\bottomrule
\end{tabular}%
}
\end{table}

\section{Training details}
\label{app:training-details}

This section details the training process of \system.

\subsection{Main Training Process}

Table~\ref{tab:training-schedule} summarizes the complete schedules.
During training, we normalize the input images to $[-1, 1]$.
When training on 512 low-resolution crops, we use the a subset of 4.8~M randomly sampled images from SAM-1B~\citep{kirillov2023sam} and Open Images~\citep{kuznetsova2020openimages}.
When training on 1024 high-resolution crops, we use 4,035 images: 800 DIV2K
images~\citep{agustsson2017div2k}, 2,650 Flickr2K
images~\citep{lim2017edsr}, and 585 CLIC training images~\citep{clic2020}.
The ROI stages use the same data as 1024 high-resolution crops, augmented with offline extracted face and text boxes. 

\begin{table}[h]
\centering
\caption{Training schedules.}
\label{tab:training-schedule}
\small
\setlength{\tabcolsep}{4.4pt}
\resizebox{\textwidth}{!}{%
\begin{tabular}{llccccl}
\toprule
Model & Stage & Crop & Steps & Batch & $\lambda$ range & Learning-rate schedule\\
\midrule
\multirow{2}{*}{MSE}
& LR training & $512$ & $500$k & 8 & $24$--$672$
& $10^{-3}$ (200k) $\to$ $10^{-4}$ (200k) $\to$ $10^{-5}$ (100k)\\
& HR training & $1024$ & $50$k & 8 & $24$--$672$
& $10^{-4}$ (40k) $\to$ $10^{-5}$ (10k)\\
\midrule
\multirow{3}{*}{Perceptual}
& Stage I LR & $512$ & $300$k & 16 & $0.5$--$12.5$
& $10^{-3}$ (200k) $\to$ $10^{-4}$ (100k)\\
& Stage I HR & $1024$ & $50$k & 16 & $0.5$--$12.5$
& $10^{-4}$ (40k) $\to$ $10^{-5}$ (10k)\\
& Stage II & $1024$ & $20$k & 32 & $3$--$25$
& $10^{-4}$ (10k) $\to$ $10^{-5}$ (10k)\\
\bottomrule
\end{tabular}}
\end{table}

\subsection{More Details of Perceptual and ROI Optimization}

\textbf{Hyperparameters.}
For perceptual optimization, we set $\phi_{\rm MSE}=5$ and define $D_{\rm LPIPS}=D_{\rm LPIPS\text{-}VGG}+0.5D_{\rm LPIPS\text{-}Alex}$.
We use $\phi_{\rm LPIPS}^{\rm I}=1$ and $\phi_{\rm LPIPS}^{\rm II}=0.25$.
In Stage II, we set $\phi_{\rm GAN}=0.005$, $\phi_{\rm ROI}=0.0125$, $\phi_{\rm FaceNet}=1$, and $\phi_{\rm CRNN}=2$.
We additionally apply ROI-restricted MSE and LPIPS supervision, each with an additional weight of $0.125$.

\textbf{GAN Training.}
For latent-conditioned adversarial training, we extract features from
blocks 3, 6, 9, and 12 of a frozen DINOv2 ViT-B/14 backbone.
At each block, the detached latent $\hat y$ is resized and projected,
then concatenated with the image features for discrimination.
We use a hinge discriminator loss and the negative mean fake-image
logit as $D_{\rm GAN}$.

\textbf{Effectiveness of Perceptual Training Designs.}
As shown in Table~\ref{tab:percetual_ablation}, we examine the effectiveness of each design.
LPIPS-first perceptual pre-training assigns a larger weight to the LPIPS loss, encouraging stronger perceptual detail synthesis, with consistent improvements in LPIPS-VGG, FID, and regional reconstruction quality.
Adversarial training with a DINO-based projected GAN loss plays a key role in improving perceptual realism: removing the GAN loss substantially degrades FID by 49.8\%.
ROI training primarily benefits perceptually important regions, with its removal causing large degradations of 37.7\% and 20.8\% in face and text PSNR, respectively, while having a relatively limited effect on the overall metrics.
These designs provide complementary benefits, and their combination achieves the best overall perceptual quality.

\begin{table}[h]
\centering
\caption{Ablation study on perceptual training designs. BD-Rate is measured.}
\label{tab:percetual_ablation}
\small
\setlength{\tabcolsep}{5pt}
\begin{tabular}{lccccc}
\toprule
Training variant & PSNR & LPIPS-VGG & FID & Face PSNR& Text PSNR\\
\midrule
Stage I w/o high LPIPS weight & -0.4\% & +6.0\% & +5.7\% & +6.1\% & +1.6\%\\
Stage II w/o GAN Loss & --3.6\% &	--6.4\%	& +49.8\% & --4.1\%	& --4.5\% \\
Stage II w/o ROI & +0.2\% & --2.4\% & --3.7\% & +37.7\% & +20.8\%\\
\midrule
Proposed perceptual optimization & 0.0\% & 0.0\% & 0.0\% & 0.0\% & 0.0\% \\
\bottomrule
\end{tabular}
\end{table}

\subsection{Meta Prior Post-Training}

After main training, we expand the factorized prior into $N=64$ banks for each QP and fit them in two stages.
All other codec parameters remain fixed during this procedure.

\textbf{Complexity-stratified initialization.}
For each hyperlatent location, we compute three luminance statistics from the corresponding input region: mean absolute gradient, local standard deviation, and coarse-scale standard deviation interpolated to the hyperlatent grid.
Each statistic is normalized by its 75th percentile over the fitting set and clipped to $[0,4]$.
Their weighted sum, with weights $(0.5,0.3,0.2)$, defines the complexity score.
We partition these locations into $N$ quantile bins, copy the pretrained QP-specific factorized prior into every bank, and fit each bank to the hyperlatent vectors in its bin.
This initializes banks specialized to different local content complexities, rather than assigning one bank to an entire image.

\textbf{Spatially adaptive hard-EM refinement.}
We replace the complexity-based assignments with position-wise minimum-cost assignments.
In detail, we adopt an expectation-maximization (EM) procedure to jointly optimize the prior models and their assignment.
The hard E-step selects
\begin{equation}
k^\star_{q,r}
= \arg\min_{k\in\{1,\ldots,N\}}
\left[-\sum_{c=1}^{Z}
\log_2 p_{q,k,c}(\hat z_{q,r,c})\right],
\end{equation}
where $p_{q,k,c}$ is the discrete probability mass of channel $c$ in bank $k$ at QP $q$.
The M-step refits each bank by minimizing its normalized cross-entropy on
the channel-wise symbol histograms of its currently assigned vectors.
Thus, different positions in the same image can use different banks,
regardless of their initial complexity bins.

We fit Meta Prior on hyperlatents extracted from $512\times512$ crops of 4,035 high-resolution images.
Fitting starts with a 500-step initialization, followed by six 300-step refinements, totaling 2,300 optimizer steps per QP.
Each fitting stage uses a learning rate of $10^{-2}$ with cosine decay to $2\times10^{-4}$. Optimization is full-batch over the aggregated hyperlatent histograms for each QP.

\section{Evaluation details}
\label{app:evaluation-details}

This section details the training process of \system.

\subsection{Time Measurement}

Latency is measured on a single CPU thread at $1088\times1920$, without hardware codec acceleration.
Neural inference uses batch size one and one inference stream.

\textbf{HEIC.}
We use libheif~1.17.6 with x265~3.5 for encoding and libde265~1.0.15 for decoding.
Encoding uses 8-bit YCbCr 4:2:0, \texttt{preset=slow}, and \texttt{tune=psnr}, with additional codec worker threads disabled.
We also evaluated YCbCr 4:4:4, but observed worse rate--distortion performance and lower coding speed, possibly reflecting less effective optimization for 4:4:4 in this implementation.
The benchmark operates on in-memory RGB inputs and bitstreams.
Disk I/O is excluded.

\textbf{HM \& VTM.}
We use HM-16.25 and VTM-17.0 reference implementations with 8-bit YUV 4:4:4 and intra-only coding.
Their timings use the native YUV interface and do not include RGB conversion.
To reduce the effect of fixed command-line initialization overhead, we measure identical 5-frame and 15-frame runs and estimating the per-picture latency as $(T_{15}-T_{5})/10$.

\textbf{Neural codecs.}
For compared neural codec, we evaluate the supported combinations of BF16, FP16, and FP32 precision with OpenVINO and PyTorch, and report the fastest measured configuration under the same single-thread
constraint.
Model loading, graph conversion and compilation, and warm-up are excluded from steady-state inference timing.
Entropy coding is included in the reported codec latency, using the same rANS implementation as in \system for consistent timing.

\subsection{Baseline Implementations}

For all methods, we use their officially released implementations, with the following exceptions.

For \textbf{HiFiC}~\citep{mentzer2020hific}, we were unable to obtain usable official pretrained models during our experiments in September 2026.
We therefore use the officially published rate--distortion measurements released with TensorFlow Compression.\footnote{\url{https://github.com/tensorflow/compression/blob/master/models/hific/data.csv}}.

For \textbf{MS-Hyperprior}~\citep{minnen2018joint}, \textbf{ELIC}~\citep{he2022elic}, and \textbf{MRIC}~\citep{agustsson2023mric}, our measurements use third-party implementations and their pretrained models.
Specifically, we use CompressAI's \texttt{mbt2018-mean} model for MS-Hyperprior, VincentChandelier's reimplementation for ELIC,\footnote{\url{https://github.com/VincentChandelier/ELiC-ReImplemetation}} and Nikolai10's TensorFlow reimplementation for MRIC.\footnote{\url{https://github.com/Nikolai10/MRIC}}

\begin{figure}[h]
    \centering
    \includegraphics[width=\textwidth]{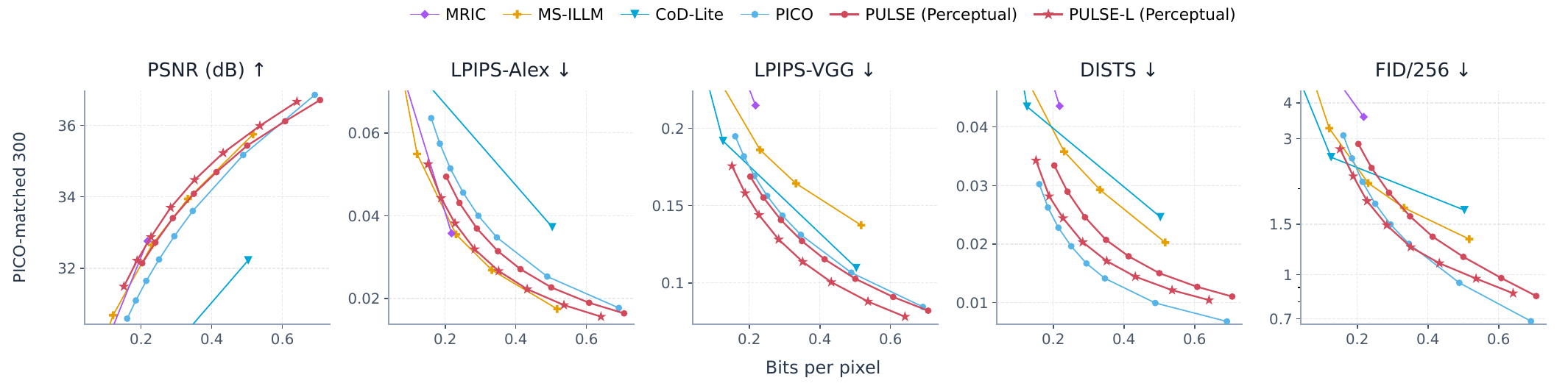}
    \caption{Rate--distortion curves on the 300-image CLIC test subset released by PICO.}
    \label{fig:rd-pico-match}
\end{figure}

For \textbf{PICO}~\citep{tatwawadi2026pico}, we reproduce their architecture based on the published specifications for complexity and latency measurements.
PICO does not release an implementation, but provides reconstructions for a 300-image subset of the CLIC 2020 test set~\citep{clic2020}.
We therefore use the same subset for comparison in Figure~\ref{fig:rd-pico-match} and Table~\ref{tab:bdrate-pico-matched-300}.
\system-L is $2.1\times$ and $2.9\times$ faster for encoding and decoding, respectively, while achieving better BD-Rate on PSNR and FID.
\label{app:pico}

\begin{table}[h]
\centering
\caption{BD-rate comparison on PICO-matched 300-image CLIC test subset. Anchor: MS-ILLM.}
\label{tab:bdrate-pico-matched-300}
\setlength{\tabcolsep}{5pt}
\renewcommand{\arraystretch}{1.10}
\resizebox{0.8\linewidth}{!}{%
\begin{tabular}{@{}lccccc@{}}
\toprule
\textbf{Method} & PSNR & LPIPS-Alex & LPIPS-VGG & DISTS & FID \\
\midrule
MRIC & -1.2\% & +9.7\% & +58.5\% & +51.1\% & +98.4\% \\
MS-ILLM & 0\% & 0\% & 0\% & 0\% & 0\% \\
CoD-Lite & +160\% & +109.4\% & -39.7\% & -1.3\% & +9.4\% \\
PICO & +18.2\% & +50.7\% & -30.6\% & -52.3\% & -6\% \\
\textbf{PULSE (Perceptual)} & \textbf{+2.2\%} & \textbf{+31.1\%} & \textbf{-38.2\%} & \textbf{-29.3\%} & \textbf{+12.7\%} \\
\textbf{PULSE-L (Perceptual)} & \textbf{-7.1\%} & \textbf{+8.9\%} & \textbf{-48.9\%} & \textbf{-45.2\%} & \textbf{-18.2\%} \\
\bottomrule
\end{tabular}}
\end{table}

\subsection{Comparison on More Benchmarks}
\label{app:rd-comparisons}

\begin{figure*}[t]
    \centering
    \includegraphics[width=\textwidth]{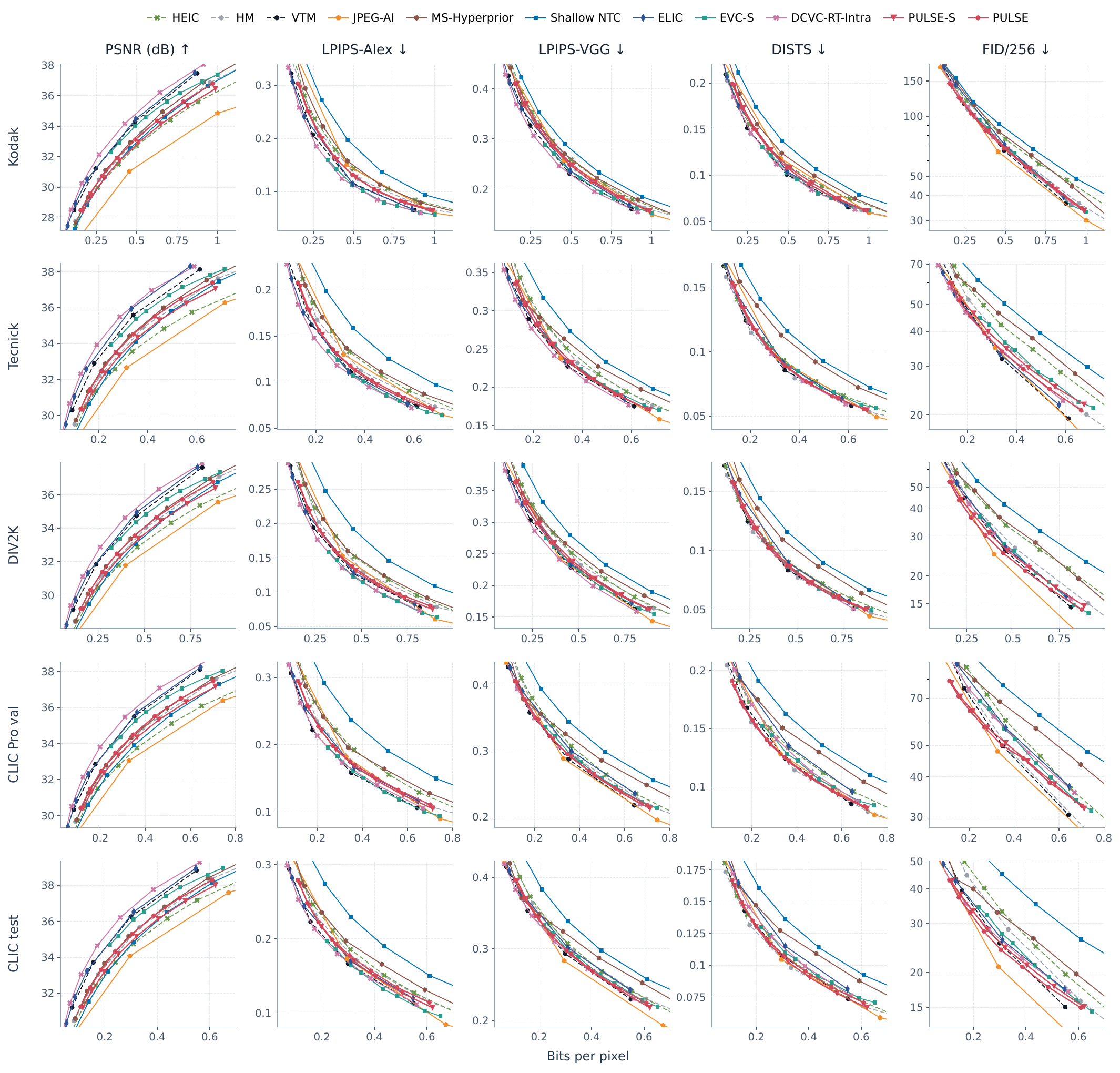}
    \caption{Rate-distortion curves on more benchmarks for MSE-optimized codecs.}
    \label{fig:rd-mse}
\end{figure*}

\begin{figure*}[t]
    \centering
    \includegraphics[width=\textwidth]{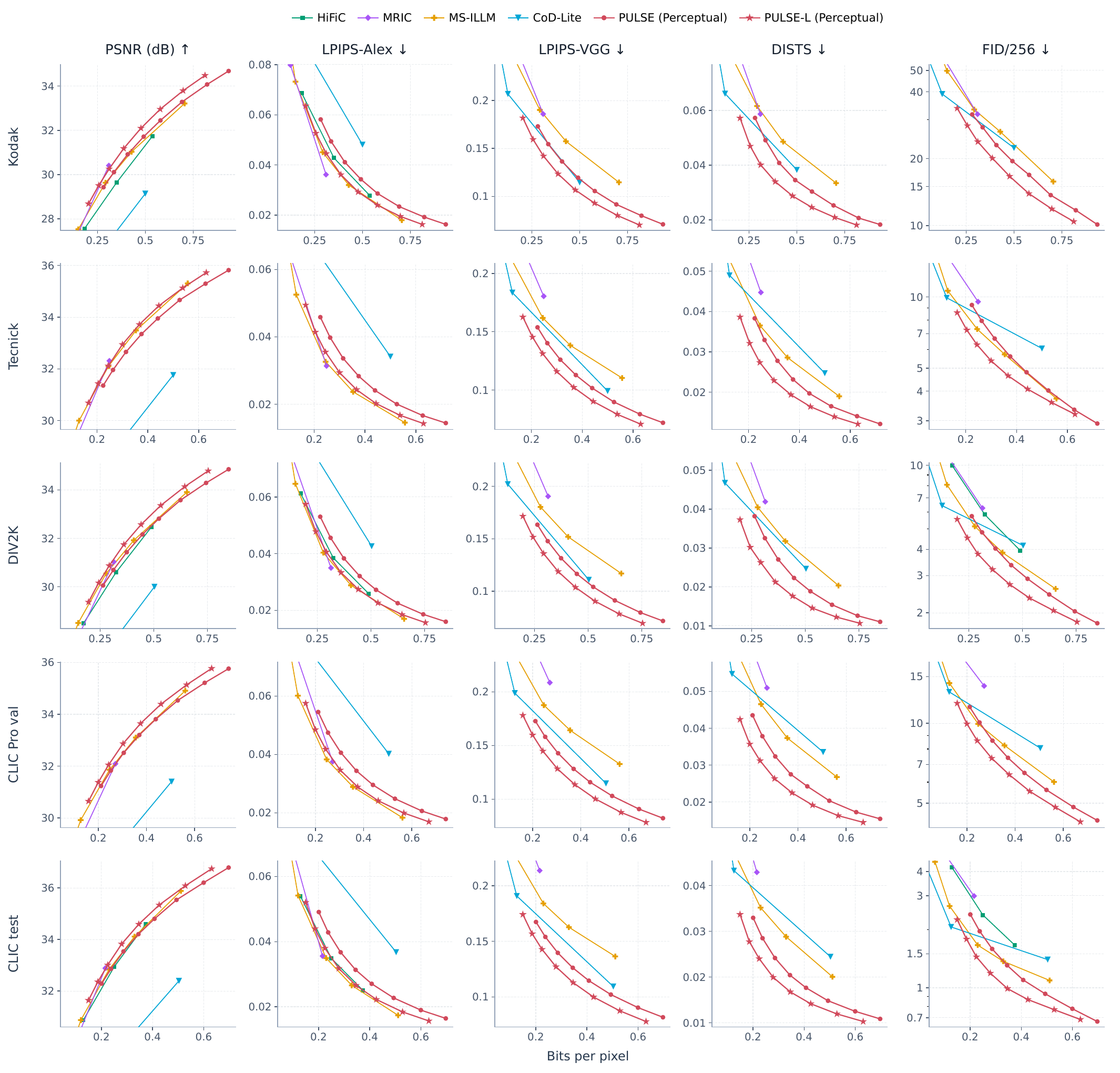}
    \caption{Rate-distortion curves on more benchmarks for perceptual-optimized codecs.}
    \label{fig:rd-perc}
\end{figure*}

In Figures~\ref{fig:rd-mse},~\ref{fig:rd-perc} and Tables~\ref{tab:bdrate-psnr}--\ref{tab:bdrate-fid}, we present rate-distortion curves and BD-Rates across more metrics and benchmarks.
We evaluate PSNR, LPIPS with VGG and AlexNet, DISTS~\citep{ding2022dists}, and FID on Kodak, Tecnick, DIV2K, CLIC Professional validation, and CLIC test sets.
We compute FID using overlapping $64\times64$ patches for Kodak and overlapping $256\times256$ patches for all other benchmarks.
Across all benchmarks and metrics, \system achieves competitive performance with substantially lower complexity.

\subsection{Additional Visual Examples}
\label{app:visual-comparisons}

\begin{figure*}[t]
    \centering
    \includegraphics[width=0.9\textwidth]{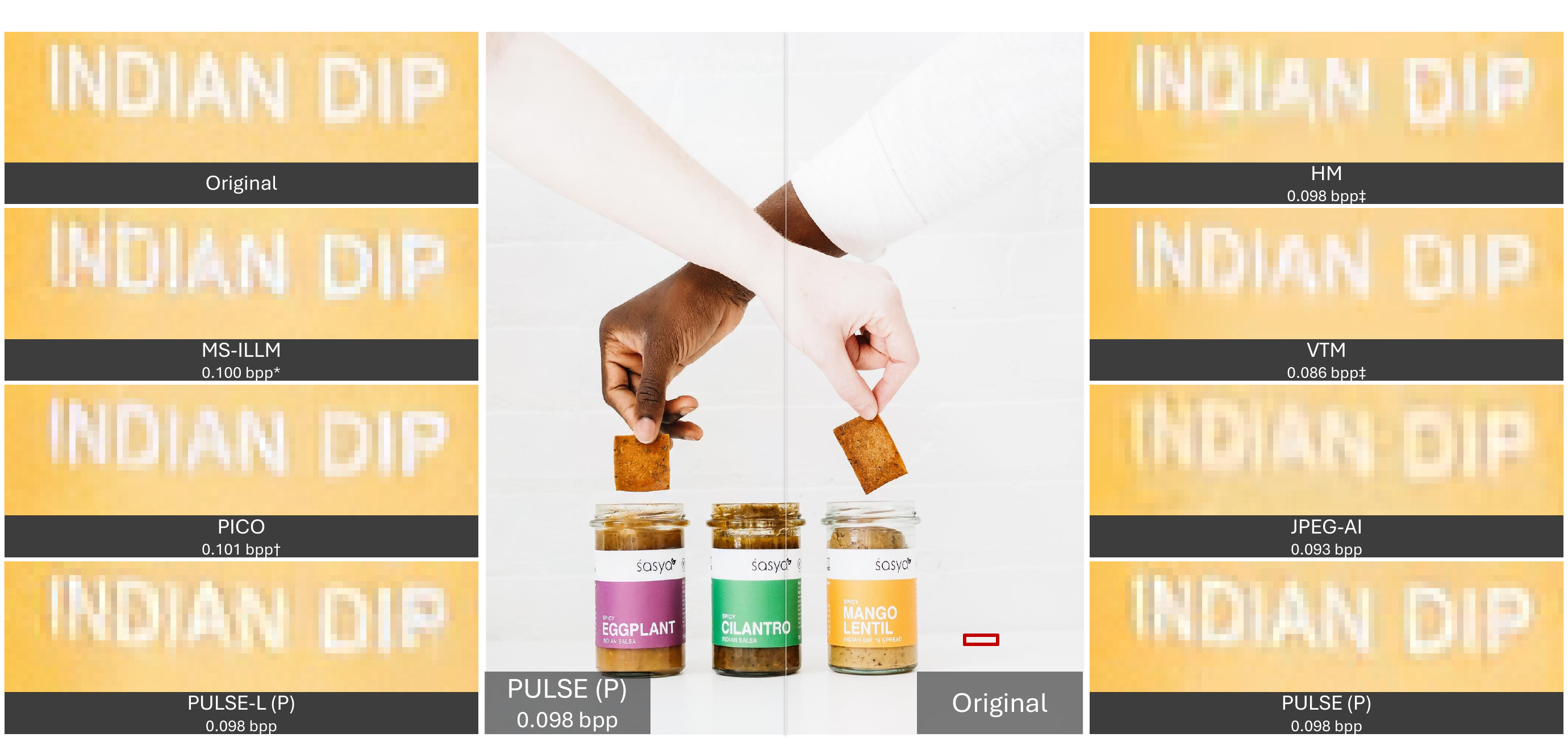}
    \includegraphics[width=0.9\textwidth]{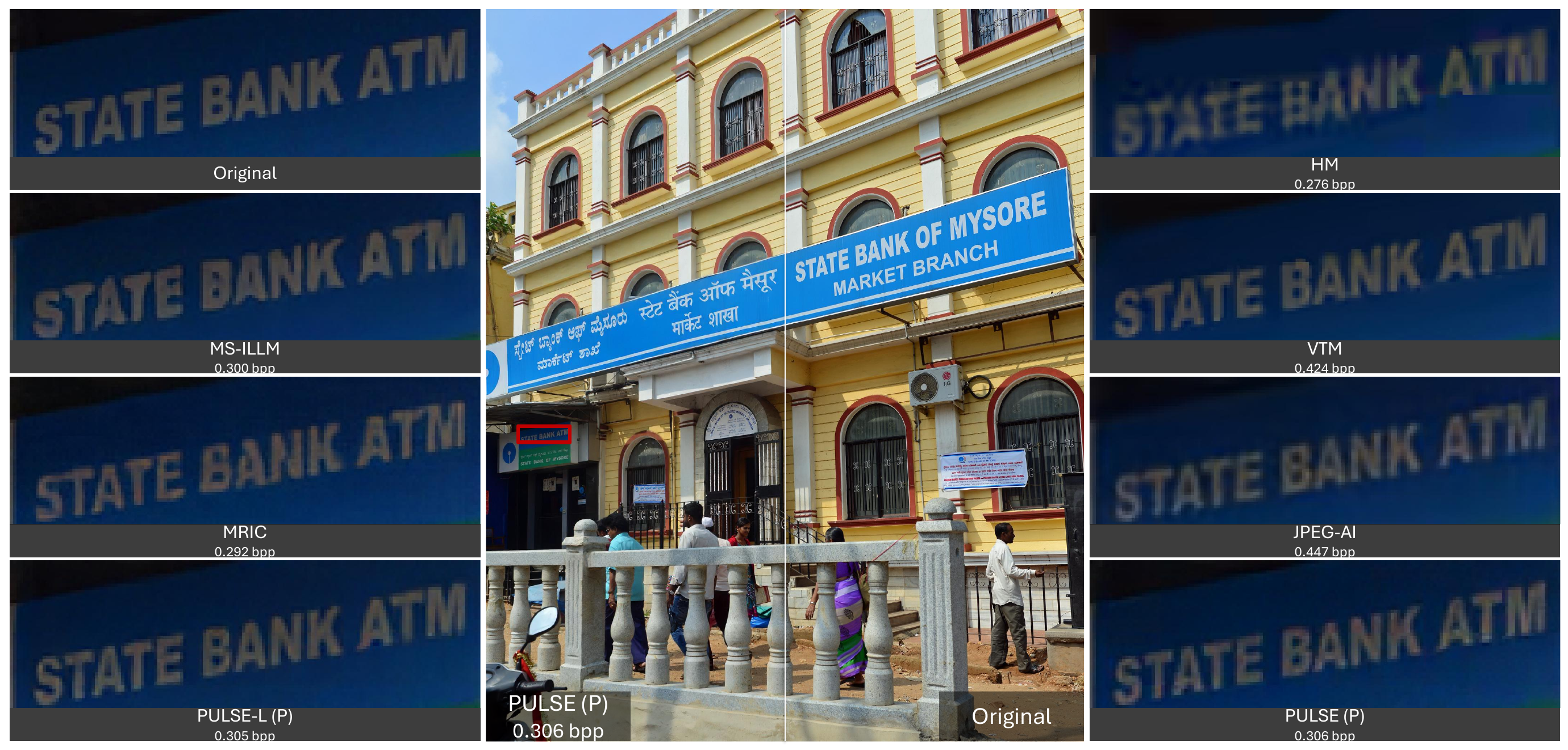}
    \caption{More visual examples on text contents.}
    \label{fig:visual_text}
\end{figure*}

\begin{figure*}[t]
    \centering
    \includegraphics[width=0.9\textwidth]{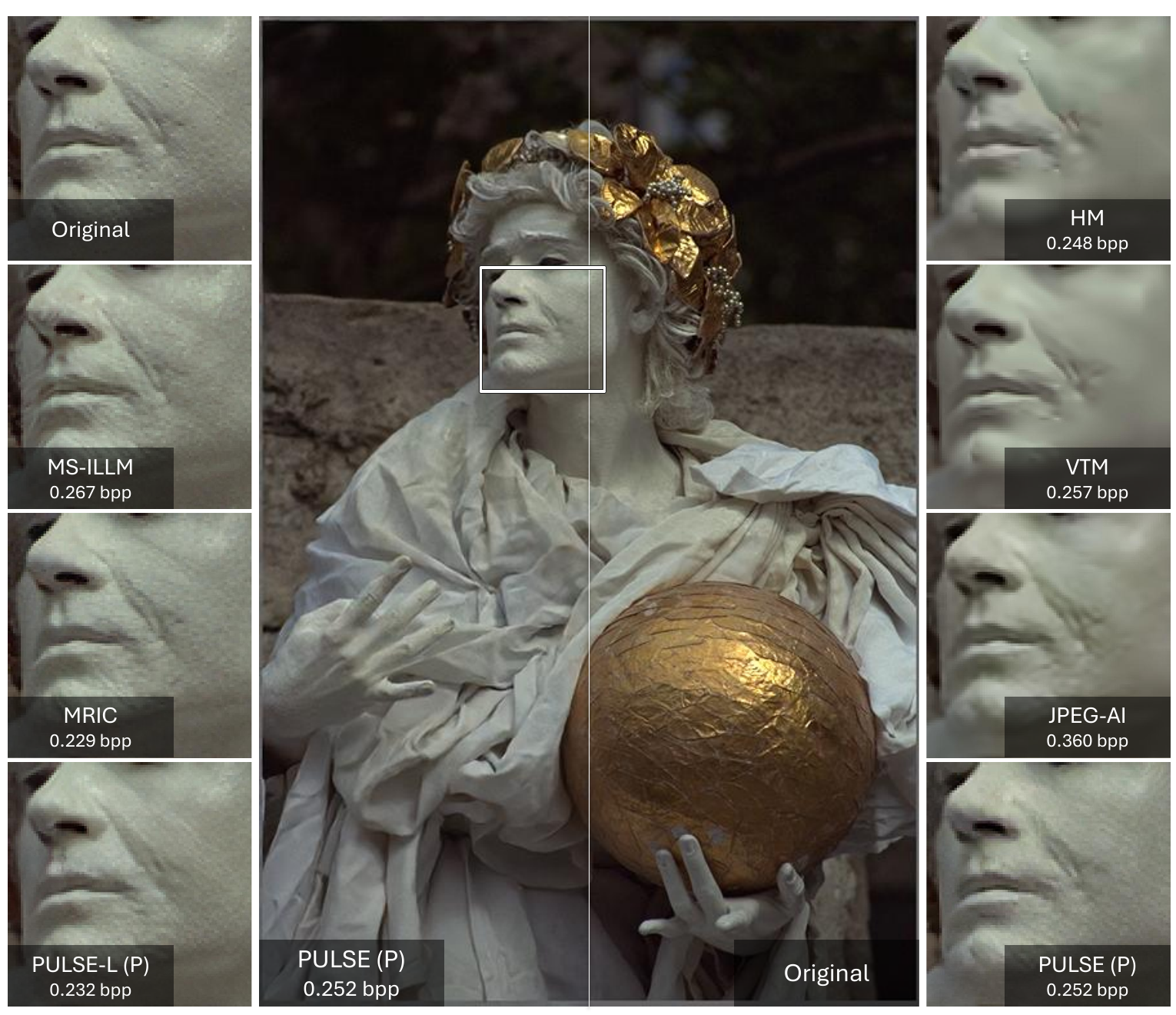}
    \includegraphics[width=0.9\textwidth]{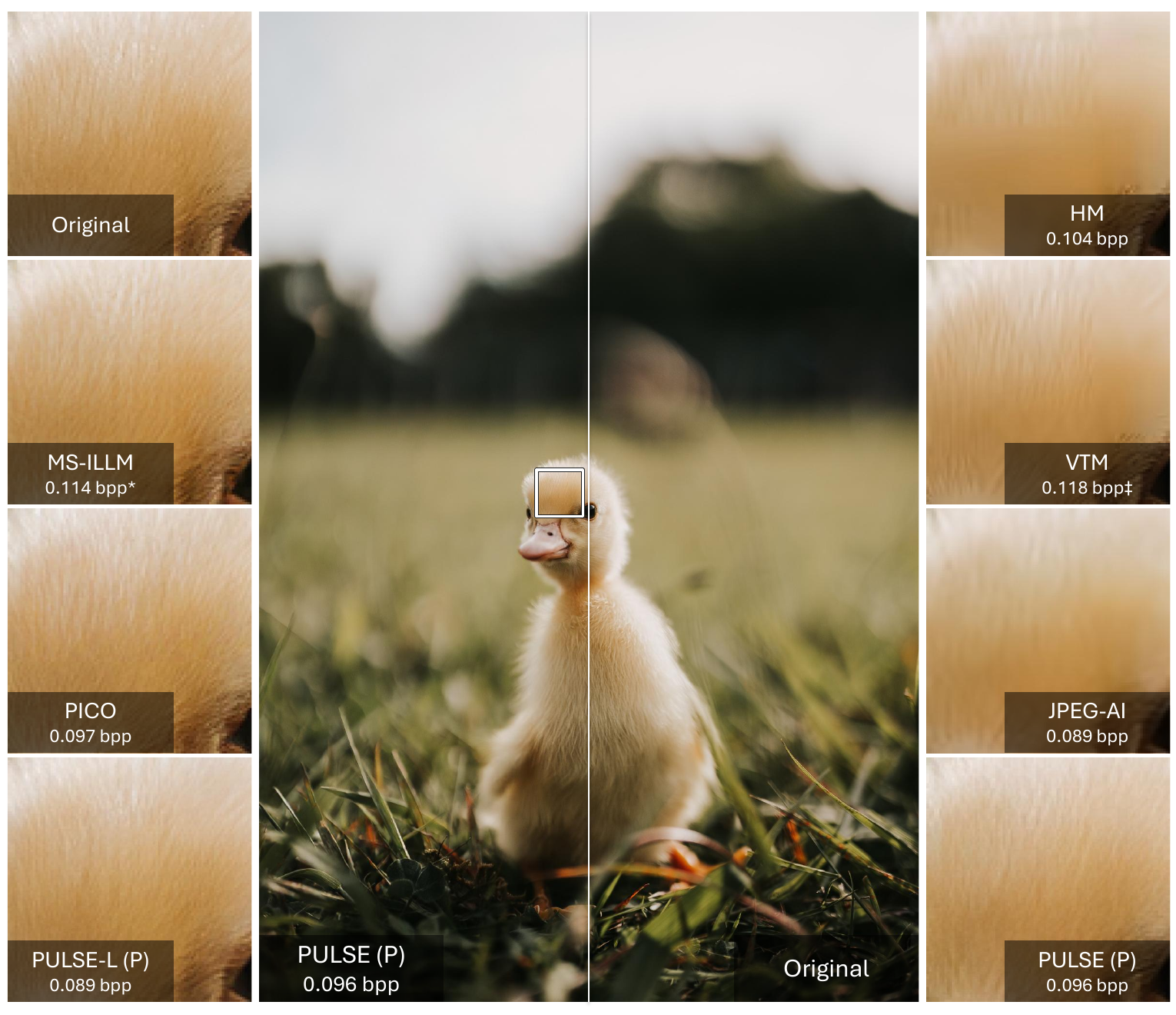}
    \caption{More visual examples on natural contents.}
    \label{fig:visual}
\end{figure*}

Figure~\ref{fig:visual_text} and Figure~\ref{fig:visual} provide additional visual comparison examples on text and natural content.
We do not include facial comparison to avoid personally identifiable information.

\section{Agentic Evolution Protocol}
\label{app:agentic-algorithm}
\label{app:agentic-details}

All E1--E7 generations follow the same human--LLM agentic research loop
described in the paper.

\begin{figure}[h]
\centering
\includegraphics[width=0.95\textwidth]{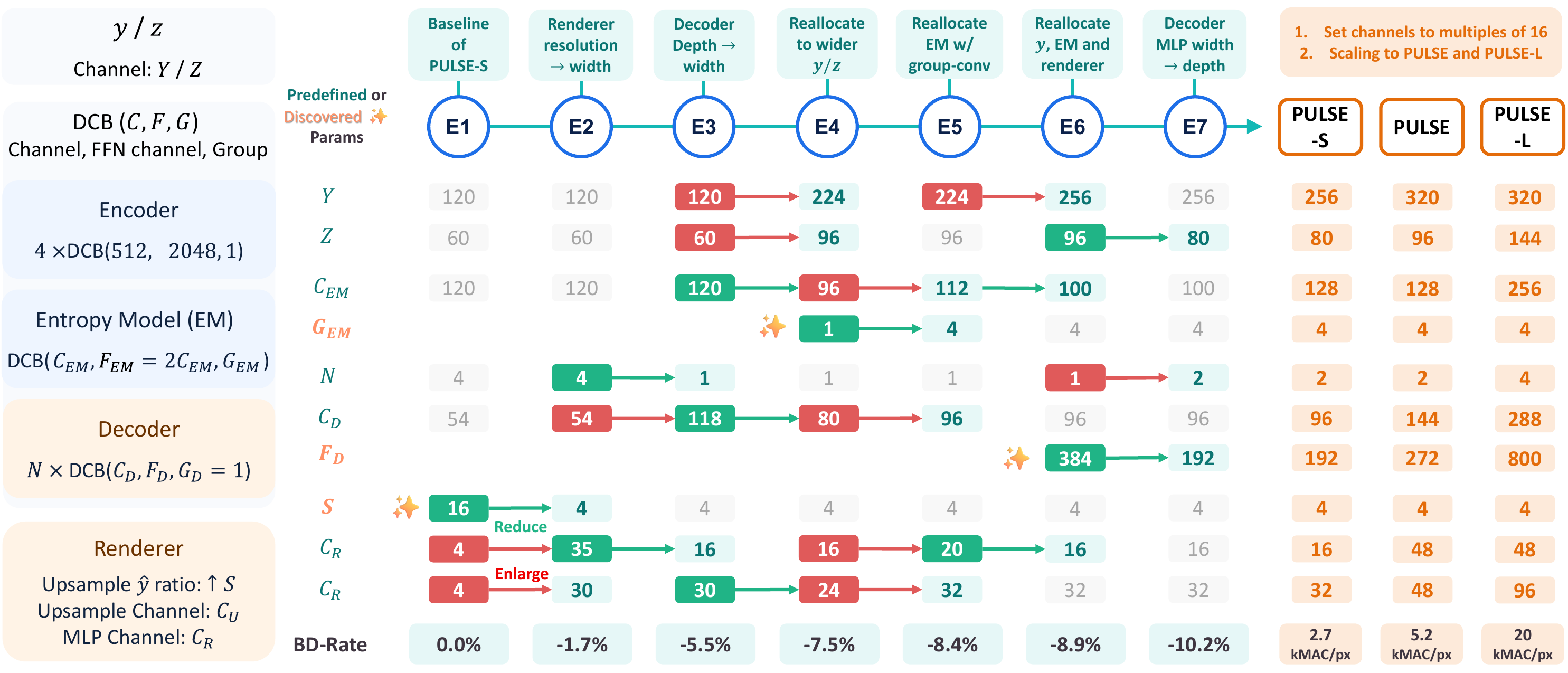}
\caption{Full agentic evolution process and final parameters at different scales.}
\label{fig:agentic_evolution_res}
\end{figure}

\begin{table}[h]
\centering
\caption{\textbf{Illustrations on each evolution.}}
\label{tab:agentic-promotions}
\small
\setlength{\tabcolsep}{5pt}
\resizebox{\textwidth}{!}{%
\begin{tabular}{lll}
\toprule
Transition & Main change & Motivation\\
\midrule
E1$\rightarrow$E2 & Move rendering to a $1/4$-resolution grid
& relieve the narrow full-resolution renderer\\
E2$\rightarrow$E3 & Replace four narrow blocks with one wider block
& concentrate low-resolution capacity\\
E3$\rightarrow$E4 & Increase main and hyper latent capacity
& address pressure in shared representations\\
E4$\rightarrow$E5 & Introduce grouped pointwise convolution
& reclaim entropy-model computation\\
E5$\rightarrow$E6 & Reallocate internal width to the main latent
& strengthen shared latent capacity\\
E6$\rightarrow$E7 & Reduce FFN channels and increase decoder depth
& improve nonlinear decoder processing\\
E7$\rightarrow$stop & Train three further candidates
& none improves E7\\
\bottomrule
\end{tabular}
}
\end{table}

\subsection{Implementation Details}

For LLM agent, we use \texttt{gpt-5.6-sol} with high reasoning effort.
The model receives the probe summaries, network graph, parameter-sharing constraints, component-level MAC accounting, and the complete history of earlier rounds.
Each interaction produces two to four candidate suggestions.
A human researcher checks their interpretation, implementation feasibility, and budget accounting, then fixes three architectures for full training in every round.

All three candidates are trained from scratch with the same data, QP range, optimizer, and schedule.
The human researcher reviews the measured rate--distortion curves and promotes the best improving candidate.
If no candidate improves the current parent, the search terminates.

\subsection{Full Evolution Trajectory}

As shown in Figure~\ref{fig:agentic_evolution_res}, the reported search contains one initial model training, seven rounds with three trained candidates per round, and several finalized model trainig at different scale. 
Six rounds promote E2 through E7.
In the final round after E7, all three candidates underperform E7, so the search stops.
Table~\ref{tab:agentic-promotions} records the changes and motivations at high level.

\begin{table*}[t]
\centering
\caption{\textbf{BD-Rate comparison on PSNR.} Anchor: HM-16.25.}
\label{tab:bdrate-psnr}
\setlength{\tabcolsep}{5pt}
\renewcommand{\arraystretch}{1.10}
\resizebox{0.8\linewidth}{!}{%
\begin{tabular}{@{}lrrrrr@{}}
\toprule
\textbf{Method} & Kodak & Tecnick & DIV2K & CLIC pro val & CLIC test \\
\midrule
HEIC & +14.3\% & +28.4\% & +21.1\% & +25.2\% & +16.9\% \\
HM-16.25 & 0\% & 0\% & 0\% & 0\% & 0\% \\
VTM-17.0 & -21.3\% & -23.8\% & -22.8\% & -23.4\% & -23\% \\
\midrule
JPEG-AI & +54.3\% & +41.2\% & +33\% & +37.4\% & +33.2\% \\
MS-Hyperprior & -2.2\% & -4.6\% & -3.3\% & -1.7\% & -2\% \\
Shallow NTC & +6.3\% & +9.9\% & +9.3\% & +13\% & +11.9\% \\
ELIC & -25.1\% & -32\% & -27.6\% & -26.2\% & -26.2\% \\
EVC-S & -12.6\% & -14.3\% & -13.5\% & -14.3\% & -13.2\% \\
DCVC-RT-Intra & -31.1\% & -36.4\% & -33.3\% & -33.2\% & -33.7\% \\
\textbf{PULSE-S} & \textbf{+9.9\%} & \textbf{+7.7\%} & \textbf{+6.5\%} & \textbf{+5.9\%} & \textbf{+9.4\%} \\
\textbf{PULSE} & \textbf{+1.2\%} & \textbf{-2.1\%} & \textbf{-2.4\%} & \textbf{-2.9\%} & \textbf{+0.3\%} \\
\midrule
HiFiC & \textcolor{gray}{+61.9\%} & \textcolor{gray}{\textemdash} & \textcolor{gray}{+37.3\%} & \textcolor{gray}{\textemdash} & \textcolor{gray}{+45.1\%} \\
MRIC & \textcolor{gray}{+22.2\%} & \textcolor{gray}{+18.3\%} & \textcolor{gray}{+22.3\%} & \textcolor{gray}{+46.1\%} & \textcolor{gray}{+29.1\%} \\
MS-ILLM & \textcolor{gray}{+39.6\%} & \textcolor{gray}{+23\%} & \textcolor{gray}{+23.3\%} & \textcolor{gray}{+38\%} & \textcolor{gray}{+38.7\%} \\
CoD-Lite & \textcolor{gray}{+167.9\%} & \textcolor{gray}{+162.2\%} & \textcolor{gray}{+145.8\%} & \textcolor{gray}{+218.8\%} & \textcolor{gray}{+232.8\%} \\
\textbf{PULSE (Perceptual)} & \textcolor{gray}{+38.8\%} & \textcolor{gray}{+42.4\%} & \textcolor{gray}{+33.7\%} & \textcolor{gray}{+45.5\%} & \textcolor{gray}{+49.6\%} \\
\textbf{PULSE-L (Perceptual)} & \textcolor{gray}{+24.4\%} & \textcolor{gray}{+26\%} & \textcolor{gray}{+19.4\%} & \textcolor{gray}{+30.8\%} & \textcolor{gray}{+34.5\%} \\
\bottomrule
\end{tabular}}
\end{table*}

\begin{table*}[t]
\centering
\caption{\textbf{BD-Rate comparison on LPIPS-Alex.} Anchor: MS-ILLM.}
\label{tab:bdrate-lpips-alex}
\setlength{\tabcolsep}{5pt}
\renewcommand{\arraystretch}{1.10}
\resizebox{0.8\linewidth}{!}{%
\begin{tabular}{@{}lrrrrr@{}}
\toprule
\textbf{Method} & Kodak & Tecnick & DIV2K & CLIC pro val & CLIC test \\
\midrule
HEIC & \textcolor{gray}{+890\%} & \textcolor{gray}{+820\%} & \textcolor{gray}{+900\%} & \textcolor{gray}{+1420\%} & \textcolor{gray}{+1330\%} \\
HM-16.25 & \textcolor{gray}{+840\%} & \textcolor{gray}{+800\%} & \textcolor{gray}{+880\%} & \textcolor{gray}{+1340\%} & \textcolor{gray}{+1450\%} \\
VTM-17.0 & \textcolor{gray}{+740\%} & \textcolor{gray}{+690\%} & \textcolor{gray}{+790\%} & \textcolor{gray}{+1140\%} & \textcolor{gray}{+1250\%} \\
\midrule
JPEG-AI & \textcolor{gray}{+830\%} & \textcolor{gray}{+700\%} & \textcolor{gray}{+760\%} & \textcolor{gray}{+1130\%} & \textcolor{gray}{+1120\%} \\
MS-Hyperprior & \textcolor{gray}{+1010\%} & \textcolor{gray}{+920\%} & \textcolor{gray}{+1000\%} & \textcolor{gray}{+1630\%} & \textcolor{gray}{+1640\%} \\
Shallow NTC & \textcolor{gray}{+1530\%} & \textcolor{gray}{+1300\%} & \textcolor{gray}{+1540\%} & \textcolor{gray}{+2410\%} & \textcolor{gray}{+2360\%} \\
ELIC & \textcolor{gray}{+770\%} & \textcolor{gray}{+600\%} & \textcolor{gray}{+760\%} & \textcolor{gray}{+1100\%} & \textcolor{gray}{+1110\%} \\
EVC-S & \textcolor{gray}{+530\%} & \textcolor{gray}{+710\%} & \textcolor{gray}{+700\%} & \textcolor{gray}{+1320\%} & \textcolor{gray}{+1460\%} \\
DCVC-RT-Intra & \textcolor{gray}{+710\%} & \textcolor{gray}{+550\%} & \textcolor{gray}{+680\%} & \textcolor{gray}{+970\%} & \textcolor{gray}{+980\%} \\
\textbf{PULSE-S} & \textcolor{gray}{+850\%} & \textcolor{gray}{+770\%} & \textcolor{gray}{+880\%} & \textcolor{gray}{+1410\%} & \textcolor{gray}{+1500\%} \\
\textbf{PULSE} & \textcolor{gray}{+860\%} & \textcolor{gray}{+770\%} & \textcolor{gray}{+890\%} & \textcolor{gray}{+1410\%} & \textcolor{gray}{+1520\%} \\
\midrule
HiFiC & +12\% & \textemdash & +9.4\% & \textemdash & +7.8\% \\
MRIC & -8.7\% & +7.7\% & +6.6\% & +23.1\% & +10.8\% \\
MS-ILLM & 0\% & 0\% & 0\% & 0\% & 0\% \\
CoD-Lite & +61.2\% & +98.7\% & +70.3\% & +96.8\% & +111.5\% \\
\textbf{PULSE (Perceptual)} & \textbf{+25.2\%} & \textbf{+31.7\%} & \textbf{+25.6\%} & \textbf{+32.2\%} & \textbf{+31.9\%} \\
\textbf{PULSE-L (Perceptual)} & \textbf{+4.9\%} & \textbf{+10.4\%} & \textbf{+5.8\%} & \textbf{+9.9\%} & \textbf{+9.4\%} \\
\bottomrule
\end{tabular}}
\end{table*}

\begin{table*}[t]
\centering
\caption{\textbf{BD-Rate comparison on LPIPS-VGG.} Anchor: MS-ILLM.}
\label{tab:bdrate-lpips-vgg}
\setlength{\tabcolsep}{5pt}
\renewcommand{\arraystretch}{1.10}
\resizebox{0.8\linewidth}{!}{%
\begin{tabular}{@{}lrrrrr@{}}
\toprule
\textbf{Method} & Kodak & Tecnick & DIV2K & CLIC pro val & CLIC test \\
\midrule
HEIC & \textcolor{gray}{+330\%} & \textcolor{gray}{+340\%} & \textcolor{gray}{+290\%} & \textcolor{gray}{+520\%} & \textcolor{gray}{+530\%} \\
HM-16.25 & \textcolor{gray}{+300\%} & \textcolor{gray}{+330\%} & \textcolor{gray}{+270\%} & \textcolor{gray}{+490\%} & \textcolor{gray}{+560\%} \\
VTM-17.0 & \textcolor{gray}{+260\%} & \textcolor{gray}{+270\%} & \textcolor{gray}{+230\%} & \textcolor{gray}{+480\%} & \textcolor{gray}{+540\%} \\
\midrule
JPEG-AI & \textcolor{gray}{+270\%} & \textcolor{gray}{+240\%} & \textcolor{gray}{+200\%} & \textcolor{gray}{+360\%} & \textcolor{gray}{+370\%} \\
MS-Hyperprior & \textcolor{gray}{+300\%} & \textcolor{gray}{+350\%} & \textcolor{gray}{+290\%} & \textcolor{gray}{+570\%} & \textcolor{gray}{+580\%} \\
Shallow NTC & \textcolor{gray}{+460\%} & \textcolor{gray}{+540\%} & \textcolor{gray}{+450\%} & \textcolor{gray}{+950\%} & \textcolor{gray}{+1000\%} \\
ELIC & \textcolor{gray}{+310\%} & \textcolor{gray}{+280\%} & \textcolor{gray}{+270\%} & \textcolor{gray}{+590\%} & \textcolor{gray}{+600\%} \\
EVC-S & \textcolor{gray}{+180\%} & \textcolor{gray}{+260\%} & \textcolor{gray}{+190\%} & \textcolor{gray}{+470\%} & \textcolor{gray}{+530\%} \\
DCVC-RT-Intra & \textcolor{gray}{+270\%} & \textcolor{gray}{+250\%} & \textcolor{gray}{+230\%} & \textcolor{gray}{+500\%} & \textcolor{gray}{+510\%} \\
\textbf{PULSE-S} & \textcolor{gray}{+310\%} & \textcolor{gray}{+300\%} & \textcolor{gray}{+270\%} & \textcolor{gray}{+500\%} & \textcolor{gray}{+570\%} \\
\textbf{PULSE} & \textcolor{gray}{+300\%} & \textcolor{gray}{+280\%} & \textcolor{gray}{+250\%} & \textcolor{gray}{+510\%} & \textcolor{gray}{+580\%} \\
\midrule
HiFiC & \textemdash & \textemdash & \textemdash & \textemdash & \textemdash \\
MRIC & +8.9\% & +37.8\% & +39.3\% & +61.1\% & +60.1\% \\
MS-ILLM & 0\% & 0\% & 0\% & 0\% & 0\% \\
CoD-Lite & -39.3\% & -26.6\% & -34.8\% & -36.9\% & -38.3\% \\
\textbf{PULSE (Perceptual)} & \textbf{-24.7\%} & \textbf{-25.3\%} & \textbf{-28.6\%} & \textbf{-35.8\%} & \textbf{-38.1\%} \\
\textbf{PULSE-L (Perceptual)} & \textbf{-38.8\%} & \textbf{-37\%} & \textbf{-41\%} & \textbf{-47.5\%} & \textbf{-48.9\%} \\
\bottomrule
\end{tabular}}
\end{table*}

\begin{table*}[t]
\centering
\caption{\textbf{BD-Rate comparison on DISTS.} Anchor: MS-ILLM.}
\label{tab:bdrate-dists}
\setlength{\tabcolsep}{5pt}
\renewcommand{\arraystretch}{1.10}
\resizebox{0.8\linewidth}{!}{%
\begin{tabular}{@{}lrrrrr@{}}
\toprule
\textbf{Method} & Kodak & Tecnick & DIV2K & CLIC pro val & CLIC test \\
\midrule
HEIC & \textcolor{gray}{+550\%} & \textcolor{gray}{+590\%} & \textcolor{gray}{+570\%} & \textcolor{gray}{+1070\%} & \textcolor{gray}{+920\%} \\
HM-16.25 & \textcolor{gray}{+480\%} & \textcolor{gray}{+560\%} & \textcolor{gray}{+500\%} & \textcolor{gray}{+960\%} & \textcolor{gray}{+960\%} \\
VTM-17.0 & \textcolor{gray}{+570\%} & \textcolor{gray}{+630\%} & \textcolor{gray}{+610\%} & \textcolor{gray}{+1160\%} & \textcolor{gray}{+1130\%} \\
\midrule
JPEG-AI & \textcolor{gray}{+560\%} & \textcolor{gray}{+550\%} & \textcolor{gray}{+540\%} & \textcolor{gray}{+910\%} & \textcolor{gray}{+860\%} \\
MS-Hyperprior & \textcolor{gray}{+680\%} & \textcolor{gray}{+890\%} & \textcolor{gray}{+790\%} & \textcolor{gray}{+1500\%} & \textcolor{gray}{+1410\%} \\
Shallow NTC & \textcolor{gray}{+980\%} & \textcolor{gray}{+1090\%} & \textcolor{gray}{+1070\%} & \textcolor{gray}{+2090\%} & \textcolor{gray}{+1920\%} \\
ELIC & \textcolor{gray}{+670\%} & \textcolor{gray}{+680\%} & \textcolor{gray}{+710\%} & \textcolor{gray}{+1290\%} & \textcolor{gray}{+1190\%} \\
EVC-S & \textcolor{gray}{+420\%} & \textcolor{gray}{+650\%} & \textcolor{gray}{+560\%} & \textcolor{gray}{+1270\%} & \textcolor{gray}{+1330\%} \\
DCVC-RT-Intra & \textcolor{gray}{+630\%} & \textcolor{gray}{+620\%} & \textcolor{gray}{+660\%} & \textcolor{gray}{+1200\%} & \textcolor{gray}{+1090\%} \\
\textbf{PULSE-S} & \textcolor{gray}{+640\%} & \textcolor{gray}{+680\%} & \textcolor{gray}{+660\%} & \textcolor{gray}{+1130\%} & \textcolor{gray}{+1160\%} \\
\textbf{PULSE} & \textcolor{gray}{+630\%} & \textcolor{gray}{+670\%} & \textcolor{gray}{+630\%} & \textcolor{gray}{+1160\%} & \textcolor{gray}{+1210\%} \\
\midrule
HiFiC & \textemdash & \textemdash & \textemdash & \textemdash & \textemdash \\
MRIC & +11.4\% & +47.5\% & +33.3\% & +40.3\% & +51.2\% \\
MS-ILLM & 0\% & 0\% & 0\% & 0\% & 0\% \\
CoD-Lite & -35.3\% & +2.6\% & -26.4\% & -7\% & +1.2\% \\
\textbf{PULSE (Perceptual)} & \textbf{-22.6\%} & \textbf{-12.5\%} & \textbf{-22.4\%} & \textbf{-28.8\%} & \textbf{-29.6\%} \\
\textbf{PULSE-L (Perceptual)} & \textbf{-43\%} & \textbf{-32.4\%} & \textbf{-42.8\%} & \textbf{-47.1\%} & \textbf{-45.6\%} \\
\bottomrule
\end{tabular}}
\end{table*}

\begin{table*}[t]
\centering
\caption{\textbf{BD-Rate comparison on FID.} Anchor: MS-ILLM.}
\label{tab:bdrate-fid}
\setlength{\tabcolsep}{5pt}
\renewcommand{\arraystretch}{1.10}
\resizebox{0.8\linewidth}{!}{%
\begin{tabular}{@{}lrrrrr@{}}
\toprule
\textbf{Method} & Kodak & Tecnick & DIV2K & CLIC pro val & CLIC test \\
\midrule
HEIC & \textcolor{gray}{+770\%} & \textcolor{gray}{+1770\%} & \textcolor{gray}{+1940\%} & \textcolor{gray}{+2680\%} & \textcolor{gray}{+4820\%} \\
HM-16.25 & \textcolor{gray}{+630\%} & \textcolor{gray}{+1630\%} & \textcolor{gray}{+1700\%} & \textcolor{gray}{+2450\%} & \textcolor{gray}{+4890\%} \\
VTM-17.0 & \textcolor{gray}{+730\%} & \textcolor{gray}{+1480\%} & \textcolor{gray}{+1700\%} & \textcolor{gray}{+2660\%} & \textcolor{gray}{+3820\%} \\
\midrule
JPEG-AI & \textcolor{gray}{+530\%} & \textcolor{gray}{+1140\%} & \textcolor{gray}{+1130\%} & \textcolor{gray}{+1720\%} & \textcolor{gray}{+2540\%} \\
MS-Hyperprior & \textcolor{gray}{+800\%} & \textcolor{gray}{+2060\%} & \textcolor{gray}{+2120\%} & \textcolor{gray}{+3830\%} & \textcolor{gray}{+4980\%} \\
Shallow NTC & \textcolor{gray}{+1010\%} & \textcolor{gray}{+2800\%} & \textcolor{gray}{+2990\%} & \textcolor{gray}{+5590\%} & \textcolor{gray}{+8750\%} \\
ELIC & \textcolor{gray}{+970\%} & \textcolor{gray}{+1360\%} & \textcolor{gray}{+1750\%} & \textcolor{gray}{+3430\%} & \textcolor{gray}{+3820\%} \\
EVC-S & \textcolor{gray}{+480\%} & \textcolor{gray}{+1820\%} & \textcolor{gray}{+1620\%} & \textcolor{gray}{+3040\%} & \textcolor{gray}{+4890\%} \\
DCVC-RT-Intra & \textcolor{gray}{+840\%} & \textcolor{gray}{+1330\%} & \textcolor{gray}{+1700\%} & \textcolor{gray}{+3200\%} & \textcolor{gray}{+3510\%} \\
\textbf{PULSE-S} & \textcolor{gray}{+630\%} & \textcolor{gray}{+1650\%} & \textcolor{gray}{+1600\%} & \textcolor{gray}{+2530\%} & \textcolor{gray}{+4010\%} \\
\textbf{PULSE} & \textcolor{gray}{+640\%} & \textcolor{gray}{+1520\%} & \textcolor{gray}{+1480\%} & \textcolor{gray}{+2420\%} & \textcolor{gray}{+3700\%} \\
\midrule
HiFiC & \textemdash & \textemdash & +42.8\% & \textemdash & +85.1\% \\
MRIC & +8.1\% & +63.8\% & +55.6\% & +101\% & +107.1\% \\
MS-ILLM & 0\% & 0\% & 0\% & 0\% & 0\% \\
CoD-Lite & -26.4\% & +17.2\% & -6.4\% & +12.9\% & +14.2\% \\
\textbf{PULSE (Perceptual)} & \textbf{-16.7\%} & \textbf{+12\%} & \textbf{-1.7\%} & \textbf{-4.1\%} & \textbf{+17.8\%} \\
\textbf{PULSE-L (Perceptual)} & \textbf{-33.3\%} & \textbf{-17.9\%} & \textbf{-27.3\%} & \textbf{-22.5\%} & \textbf{-17.2\%} \\
\bottomrule
\end{tabular}}
\end{table*}

